\documentclass[sigconf,nonacm,natbib=true]{acmart}
\AtBeginDocument{%
  }

\acmISBN{978-1-4503-XXXX-X/2018/06}

\usepackage{color,tabularray, subcaption}

\begin{document}

\title{Look It Up: Analysing Internal Web Search Capabilities of Modern LLMs}


\author{Sahil Kale}
\email{sahilrkale05@gmail.com}
\affiliation{%
  \country{Pune, India}
}








\renewcommand{\shortauthors}{Kale, S.}

\begin{abstract}
Modern large language models increasingly integrate internal web-based retrieval to provide real-time answers, yet it remains unclear how effectively these systems identify information need, trigger retrieval, and use retrieved evidence. To understand these parameters better, we evaluate the necessity and effectiveness of internal web search through an external lens in closed-source LLMs, without access to model parameters or internal configuration. Our evaluation method comprises a static split of 783 temporally anchored factual queries answerable from pre-cutoff knowledge, designed to test whether retrieval is invoked based on factual uncertainty, and a dynamic split of 288 post-cutoff queries that require up-to-date information, designed to evaluate retrieval effectiveness under unavoidable information need. We experiment across four models across two model families and scales. Across models, enabling retrieval yields substantial accuracy gains on the static split, but systematically degrades confidence calibration. On the dynamic split, models frequently invoke retrieval yet remain below 70 percent accuracy, with failures dominated by query formulation and source selection errors rather than integration of retrieved content. We further analyse retrieval utility through outcome transitions, confidence-gated selective retrieval, cost-normalised improvement metrics, and a comparative external web-search baseline. While retrieval is inexpensive to invoke and can be selectively beneficial, repeated retrieval attempts rarely recover from early failures, and confidence becomes inflated once retrieval is available. Overall, internal web-based retrieval functions effectively as a low-latency verification mechanism, but falls short as a reliable IR pipeline, highlighting the need for improved retrieval triggering, query formulation, and evidence-aware confidence calibration in web-enabled LLMs. 
\end{abstract}




\maketitle

\section{Introduction}
The rapid adoption of large language models has expanded their use across a wide range of applications \cite{li-etal-2024-fundamental,kale2024faqgen,ke-etal-2025-adaptation}. However, these models remain fundamentally constrained by static parametric knowledge and lack direct access to up-to-date information, leading to potentially hallucinated outputs \cite{wang-etal-2025-bring,yeginbergen-etal-2025-dynamic,kale2025miragemasterymemorizationtricks,kale2025liemeknowledgegraphs}. As a result, from an information retrieval perspective, their effectiveness degrades on queries whose relevance depends on recent or evolving facts, a setting that is central to many real-world tasks \cite{liu-etal-2025-real,ko-etal-2024-growover}. Although pipeline-based external support mechanisms \cite{xie2024_weknowrag,kale2025knowrlteachinglanguagemodels} and RAG-style scaffolding \cite{qin2023queryrewriting} have resolved real-time context fetching to some extent, the integration of real-time web search tools into the interfaces and APIs of frontier LLMs by OpenAI \cite{openai_web_search_tool} and Anthropic \cite{anthropic_claude_web_search} has fundamentally reshaped the landscape of web-augmented LLMs across multiple applications. This integration embeds a retrieval decision and query formulation step within the model, enabling it to access fresh web content without external orchestration. In principle, LLMs can now be viewed as conditional Information Retrieval (IR) agents that must decide not only what to answer, but also whether retrieval is necessary and how to fetch relevant evidence.

\begin{figure}[t]
    \centering
    \includegraphics[width=0.75\columnwidth]{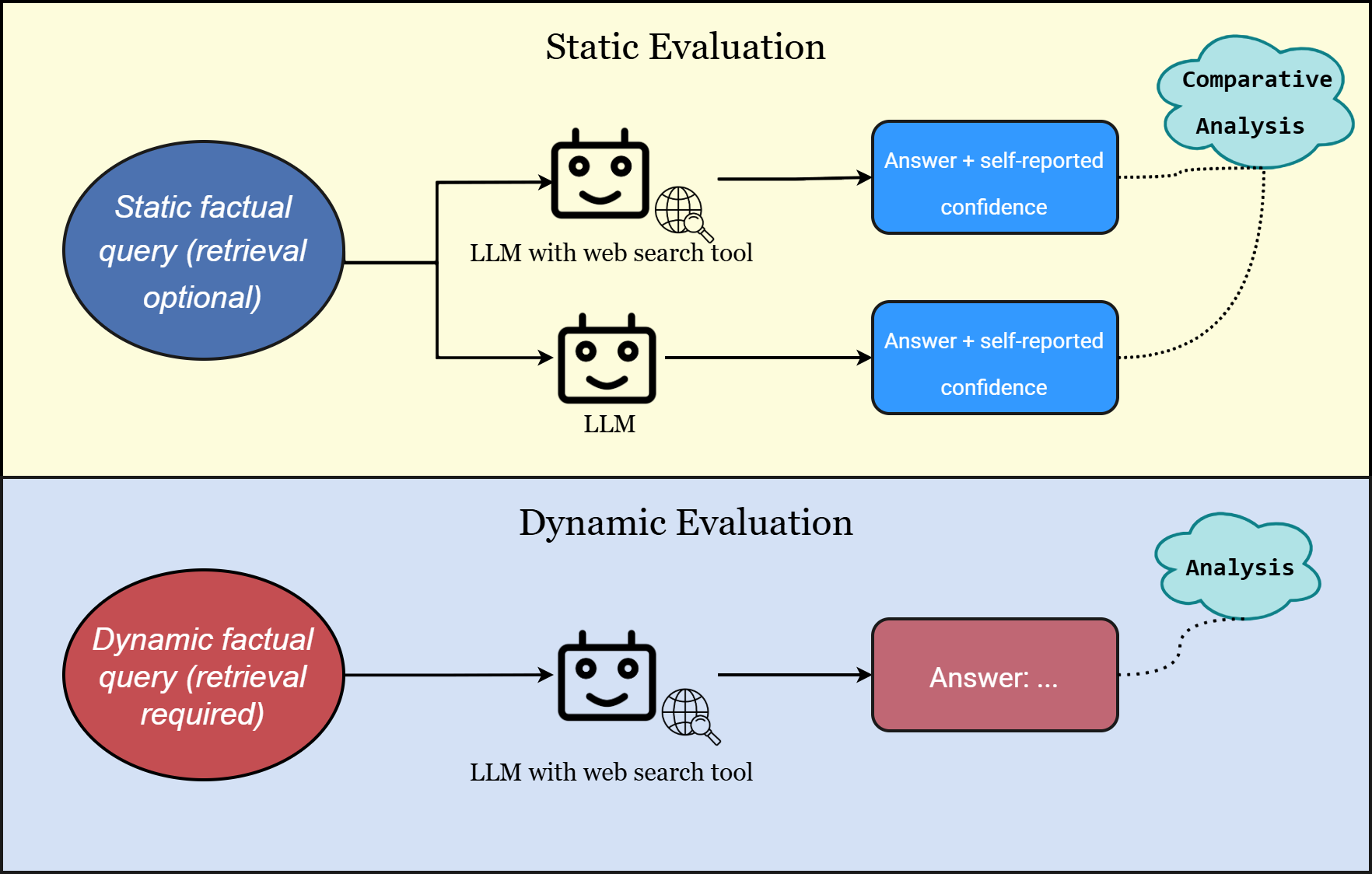}
    \caption{Overview of the evaluation framework for internal web-based retrieval in LLMs.}
    \Description{Evaluation pipeline for assessing internal web search behaviour in LLMs}
    \label{fig:eval}
\end{figure}

\begin{table*}[t]
\centering
\caption{Dataset statistics for static and dynamic evaluation splits}
\label{tab:dataset_stats}
\resizebox{\textwidth}{!}{
\begin{tblr}{
  cells = {c},
  colspec = {
    p{0.08\textwidth}   
    p{0.10\textwidth}   
    p{0.20\textwidth}   
    p{0.13\textwidth}   
    p{0.12\textwidth}   
    p{0.13\textwidth}   
    p{0.13\textwidth}   
    p{0.11\textwidth}   
  },
  hline{1} = {1-8}{},
  hline{2,4} = {-}{}
}
Split & Total queries & Source & Explicit temporal reference & Key date constraint & Average question length & Average answer length & Validation type \\
Static  & 783           & TempRAGEval filtered                  & Yes & Before May 2024     & 12.47 ± 2.76 words      & 2.13 ± 0.98 words     & Exact match        \\
Dynamic & 288           & Manually reformulated from static set & No  & After Feb 2025      & 8.68 ± 2.65 words       & 2.22 ± 1.43 words     & Exact match        
\end{tblr}
}
\end{table*}

Recent work has studied browsing agents and tool-augmented language models, largely focusing on coverage, navigation ability, and the accessibility of online sources \cite{wei2025browsecompsimplechallengingbenchmark}. From an information retrieval perspective, however, a more fundamental question remains unexplored: whether LLMs can act as efficient selective retrievers for web-based content. Specifically, it is unclear whether models can reliably decide when retrieval is necessary for answering a query, and whether they can issue efficient retrieval queries over the internet that improve answer quality without incurring unnecessary cost. In contrast to standard IR settings, where retrieval is assumed to be always on, internal web search introduces a conditional decision that must balance accuracy, latency, and resource usage.

In open-source models, such retrieval policies can be examined through internal activations or explicit tool-selection scores. Closed-source systems, however, expose only observable behaviour, requiring external evaluation to infer retrieval decision quality. We therefore focus on widely deployed commercial models from OpenAI and Anthropic, spanning both smaller and frontier model variants, as they are representative of real-world systems and increasingly used in production settings. Our approach adopts an externally grounded evaluation methodology that treats retrieval invocation and retrieval effectiveness as the main observable outcomes.

Frontier models from OpenAI and Anthropic now ship with built-in web search tools that allow models to autonomously issue retrieval requests during inference \cite{openai_web_search_tool,anthropic_claude_web_search}. Documentation from model providers suggests that LLMs internally assess whether retrieval would improve response quality, effectively implementing an implicit retrieval policy. In practice, such policies are likely driven by signals such as perceived answer uncertainty \cite{xu-etal-2025-alignment} and detection of queries whose relevance depends on post-training information \cite{cheng2024dateddatatracingknowledge}. Despite the central role of these mechanisms, there is no standardised evaluation that measures how accurately, selectively, or cost-effectively closed-source LLMs invoke web search for information retrieval.

To address this gap, we provide a reproducible evaluation protocol and versioned benchmark snapshots for studying web-enabled LLM retrieval. The benchmark examines when retrieval is necessary, when models tend to prefer it, and how effectively retrieved information contributes to correct answers on simple factual queries. The benchmark is divided into a static and a dynamic part.

The static split evaluates model behaviour with and without access to web search. It contains questions about fixed facts from before each model's knowledge cut-off, allowing a controlled assessment of retrieval strategies in terms of accuracy gains, cost-accuracy trade-off, and confidence calibration \cite{qiu-etal-2024-large}. The dynamic split evaluates retrieval-necessary cases, focusing on time-sensitive queries whose correct answers lie beyond the model’s knowledge cutoff. These queries are framed using present-tense cues such as “currently” or “as of now” to test whether models recognise the need for retrieval and issue effective search requests \cite{cheng2025dailyqabenchmarkevaluateweb}. To support sustained applicability, we additionally provide a lightweight script that periodically refreshes reference sources and answers in the dynamic split using a grounded RAG pipeline restricted to retrieval and update operations.

Overall, our paper presents a systematic and externally measurable framework for analysing information retrieval behaviour in modern LLMs equipped with web search tools. The main contributions of our work are as follows:

\begin{itemize}
    \item We introduce a versioned benchmark and evaluation protocol for black-box web-enabled LLMs that evaluates the selectivity and calibration of internal web search without relying on internal configuration
    \item We analyse LLMs' information retrieval capabilities for web content through a comparative evaluation of the benefits and cost with and without web search, as well as a stand-alone assessment of calibration in search-enabled settings
    \item We derive practical insights into selective retrieval strategies for LLM-based systems, mapping model behaviour and cost–accuracy dynamics to concrete deployment and application scenarios
\end{itemize}

\section{Related Work}

\subsection{Information Retrieval and LLMs}
LLMs are increasingly integrated into information retrieval systems as re-rankers, query rewriters, and core components of Retrieval-Augmented Generation (RAG) pipelines \cite{Dai2024CocktailAC, Wang2024BenchmarkingLL}. Recent advances include prompt-augmented and zero-shot retrieval methods such as LameR \cite{Shen2023LLMsAS}, which leverage LLM context to generate richer queries, and representation-based approaches like LMORT \cite{Sun2024LMORTLL}, which tune internal layers for retrieval metrics including pairwise similarity and vector uniformity. Iterative retrieval-verification frameworks \cite{Li2023LLatrievalLL} further position LLMs as active agents that verify evidence and generate new queries to resolve knowledge gaps.

While established benchmarks such as BEIR and MS MARCO \cite{Thakur2021BEIRAH} remain central, newer evaluations target limitations of LLM-based retrieval. Cocktail \cite{Dai2024CocktailAC} measures issues such as source bias toward LLM-generated content, and the RAG benchmark \cite{Wang2024BenchmarkingLL} evaluates robustness to noise and the ability to detect knowledge gaps. However, most prior work assumes retrieval as an explicit and fixed operation, with the LLM acting within or downstream of a search component. As a result, existing IR and RAG benchmarks do not analyse systems where retrieval is conditional, model-initiated, and dynamically invoked during inference, nor do they examine how such behaviour varies across model scales. Our work complements prior research by shifting focus from retrieval quality in fixed pipelines to retrieval decision-making within autonomous, search-enabled language models.

\subsection{Web-augmented LLMs}
Recent work has explored equipping LLMs with direct access to web content to overcome the limits of static parametric knowledge. Systems such as RaDA \cite{kim-etal-2024-rada} and DeepResearcher \cite{zheng2025deepresearcherscalingdeepresearch} embed web retrieval and browsing into agentic workflows, enabling evidence gathering during reasoning. In parallel, several benchmarks evaluate web interaction across diverse tasks. BrowserGym \cite{dechezelles2025browsergymecosystemwebagent} provides a standardised environment for web agents, Search Arena \cite{miroyan2025searcharenaanalyzingsearchaugmented} analyses multi-turn search in conversational settings, and DailyQA \cite{cheng2025dailyqabenchmarkevaluateweb} targets question answering over rapidly changing facts.

Other benchmarks examine more complex navigation and multi-step search. Mind2Web \cite{mind2web} and its extensions \cite{gou2025mind2web2evaluatingagentic} evaluate real-world task completion on live websites, WebWalkerQA \cite{wu-etal-2025-webwalker} studies structured navigation over web graphs, and BrowseComp \cite{wei2025browsecompsimplechallengingbenchmark} and its multilingual variants \cite{zhou2025browsecompzhbenchmarkingwebbrowsing} focus on persistent multi-hop search. While these benchmarks provide insight into browsing depth and execution, they typically assume retrieval or navigation is always initiated, and therefore do not examine whether or why a model chooses to retrieve. They also do not explicitly address the maintenance of time-sensitive benchmarks as underlying web content evolves.

More broadly, benchmarks on tool use in agentic workflows powered by LLMs, including Agent-SafetyBench \cite{zhang2025agentsafetybenchevaluatingsafetyllm}, AgentHarm \cite{andriushchenko2025agentharmbenchmarkmeasuringharmfulness}, T-Eval \cite{t-eval-chen-etal-2024-eval}, ACEBench \cite{chen-etal-2025-acebench}, and StableToolBench \cite{guo-etal-2024-stabletoolbench}, evaluate tool competence, safety, and execution fidelity. However, they do not directly assess retrieval selectivity, efficiency, or cost-aware invocation of web search. In contrast, our work focuses on internal web search as a retrieval policy embedded within closed-source LLMs. We treat retrieval as a conditional action rather than a fixed pipeline component, and evaluate how effectively models decide when to retrieve, how retrieval improves answer quality, and how these decisions trade off accuracy and cost.

\section{Methodology}
To capture both the calibration and the precision of LLMs in using web search, we evaluate retrieval under different information-need conditions. Figure \ref{fig:eval} illustrates the overall evaluation pipeline. Exact statistics for both splits of the final dataset are shown in Table \ref{tab:dataset_stats}. Throughout our analysis, a web search invocation denotes a single outbound retrieval request issued by the model via the provider’s built-in web search tool, independent of how many pages are accessed internally. We evaluate retrieval solely through end-to-end utility, measured by answer correctness, selective invocation behaviour, and observed failure modes. Classical ranking metrics such as nDCG@k are not reported, since internal web search does not expose ranked result lists or the combination of results used to formulate answers. We provide reproducibility of the benchmark artifact and evaluation protocol, while acknowledging that live-web retrieval itself is not exactly temporally repeatable.

\subsection{Static Split}
\subsubsection{Dataset}
The static split is constructed to evaluate retrieval selectivity in cases where external information is not strictly necessary. This setting allows us to examine how a model’s internal assessment and confidence in its parametric knowledge influences the decision to initiate retrieval, and whether retrieval is selectively invoked only when it is expected to improve answer quality. The dataset consists of precise, fully specified factual queries with a single verifiable answer prior to the model’s knowledge cutoff. 

To construct this split, we manually filter the TempRAGEval dataset \cite{zhang-etal-2025-mrag}, retaining only questions whose temporal context precedes the knowledge cutoffs of the evaluated models. This ensures that retrieval is not required for correctness, allowing a controlled analysis of unnecessary or redundant retrieval. The dataset construction proceeds as follows.

\begin{enumerate}
  \item \textit{Base corpus:} We start from the TempRAGEval test set (a temporally-perturbed QA benchmark that repurposes TimeQA \cite{chen2021timeqa} and SituatedQA \cite{zhang2021situatedqa}) with a total of 1.24k rows.
  
  \item \textit{Temporal and factual filtering:} We retain only queries that (a) contain at least one explicit temporal reference or anchor (for example, explicit years, relative phrases such as "in 2010", or constructions that indicate the fact was true at a prior date), and (b) refer to facts whose temporal context (key time in the original dataset) is before the earliest knowledge cutoff among the evaluated models. Static rules based on the 'temporal relation' and 'key date' fields of the original dataset are used for this process. This avoids ambiguity when comparing models with different cutoffs and ensures that all queries are answerable without retrieval.

  \item \textit{Manual validation and de-duplication:} The remaining queries are manually reviewed using predefined inclusion criteria. We remove examples that are duplicated, ambiguous, underspecified or require multi-step reasoning beyond a factual lookup. For each retained query, we record the canonical answer, validated semantic answer variants from the original datasets, temporal key date, and one or more supporting sources.
\end{enumerate}

Details about manual curation, decision rules and semantic variant discovery are provided in Appendix~\ref{app:static-curation}. Through this process, we obtain and release the final static split containing 783 verified factual queries. 

\subsubsection{Evaluation and Metrics}
The static split is evaluated as a controlled test of retrieval selectivity and retrieval effectiveness under conditions where external retrieval is not strictly required. Each query is presented to the same model under two conditions: once with web search disabled and once with web search enabled, with a maximum of two retrieval calls permitted. In both settings, the model is asked to report a self-assessed confidence score in the range $[0,1]$, following the setup in Appendix \ref{app:eval-prompts}. In the search-enabled condition, the model receives only the query text, without any explicit cue or instruction to invoke retrieval. This design ensures that retrieval decisions are driven by the model’s internal assessment of information need rather than prompt-level encouragement. Model responses are evaluated using normalised exact matching against the canonical answer and validated semantic answer variants provided with each query. The aim is straightforward: the model should rely on its internal factual confidence and use web search to verify or find responses to queries where it believes it has lower confidence or does not know the answer. This comparative setup allows us to measure:
\begin{enumerate}
    \item \textit{Answer accuracy and confidence calibration} under retrieval-disabled and retrieval-enabled conditions.
    \item \textit{Retrieval selectivity}, captured through the alignment between retrieval invocation decisions and the model’s stated confidence.
    \item \textit{Cost–utility trade-off}, measuring the extent to which accuracy gains justify the additional cost incurred by web search.
\end{enumerate}

\begin{table}[t]
\centering
\caption{End-to-end answer accuracy (with 95\% CI) under retrieval-disabled and retrieval-enabled settings on static and dynamic splits}
  \label{tab:acc}
\resizebox{\columnwidth}{!}{
\begin{tblr}{
  cells = {c},
  cell{1}{1} = {r=2}{},
  cell{1}{2} = {c=2}{},
  hline{1,3,7} = {-}{},
  hline{2} = {2-4}{},
  colspec = {c p{0.22\columnwidth} p{0.22\columnwidth} p{0.22\columnwidth}}
}
Model            & Static Split            &                          & Dynamic Split            \\
                 & \textit{w/o web search} & \textit{with web search} & \textit{with web search} \\
GPT-5-mini     & 0.523~± 0.027           & 0.846~± 0.035            & 0.684 ± 0.054            \\
Claude Haiku 4.5 & 0.414 ± 0.035           & 0.747 ± 0.030            & 0.608 ± 0.056           \\
GPT-5     & 0.755~± 0.049           & 0.907~± 0.032            & 0.714 ± 0.053            \\
Claude Sonnet 4.6     & 0.731~± 0.048           & 0.868~± 0.042            & 0.662 ± 0.051            
\end{tblr}
}
\end{table}

\begin{table}[t]
\centering
\caption{Accuracy comparison between internal web-based IR and an external Google Search pipeline on the dynamic split}
\label{tab:external-baseline}
\resizebox{\columnwidth}{!}{
\begin{tblr}{
  cells = {c},
  colspec = {p{0.30\columnwidth} p{0.40\columnwidth} p{0.40\columnwidth}},
  hline{1,2,6} = {-}{},
}
Model & Internal Web Search & External Google IR Pipeline \\
GPT-5-mini       & $0.684 \pm 0.054$ & $0.878 \pm 0.061$ \\
Claude Haiku 4.5 & $0.608 \pm 0.056$ & $0.853 \pm 0.065$ \\
GPT-5 & $0.714 \pm 0.053$ & $0.909 \pm 0.071$ \\
Claude Sonnet 4.6 & $0.662 \pm 0.051$ & $0.881 \pm 0.067$ \\
\end{tblr}
}
\end{table}

\subsection{Dynamic Split}
\subsubsection{Dataset}
The dynamic split evaluates retrieval necessity by testing whether models recognise time-sensitive information needs and correctly initiate web search when parametric knowledge is insufficient. Unlike the static split, all queries in this setting require external retrieval for correctness. The two splits are designed for complementary evaluation objectives and are therefore not intended as directly comparable datasets. The dataset is constructed by adapting a subset of static queries into present-tense, time-sensitive questions whose answers changed after the model’s knowledge cutoff.

Each dynamic query is designed such that parametric knowledge alone cannot yield the correct answer. The queries target well-documented recent factual changes for which the updated answer can be established from authoritative web sources. For example, a static query such as “Who was the CEO of Company X in 2021” is rewritten as “Who is the current CEO of Company X”, where the correct answer has changed after training cutoff. The construction process is as follows:

\begin{enumerate}
\item \textit{Source pool:} We start from the finalised static split of 783 examples and select candidates where (a) the answer concerns a role, office, leadership position, or other attribute likely to change over time, and (b) the question can be rewritten into a present-tense form without semantic loss.

\item \textit{Temporally neutral rewriting:} Selected queries are rewritten to express a current information need by removing historical temporal anchors and using present-tense formulations. The temporal reference is defined by the
evaluation snapshot date rather than by a date explicitly supplied in the query.

\item \textit{Post-cutoff check and validation:} We retain queries whose ground-truth answers changed after the latest knowledge cutoff among the evaluated models, verified manually using high-quality sources such as Wikipedia revision histories and reputable news outlets. We also perform validation to remove duplicated, ambiguous, underspecified or multi-step queries. For each question, gold-standard sources are recorded, and all valid semantic variants are included as reference answers.
\end{enumerate}

More details about manual efforts and semantic variant addition are provided in Appendix~\ref{app:dynamic-curation}.

Following this process, we obtain a dynamic split of 288 queries. This construction creates an evaluation setting in which external retrieval is necessary to obtain the updated answer, while the target answers are designed to remain accessible through a small number of targeted searches.  All answers were verified as correct as of 23 August 2026, and all experiments were conducted between 15 and 20 August 2026 to avoid temporal drift. The dynamic split is inherently time-sensitive and is therefore released as a versioned benchmark snapshot rather than as a permanently fixed test set. To support longer term validity, we additionally provide a lightweight script along with our benchmark that periodically refreshes reference sources and corresponding answers to remain aligned with the current day. This script uses a simple grounded RAG pipeline restricted to retrieval and update operations, ensuring that the split can be maintained without altering its original construction principles.

\subsubsection{Evaluation and Metrics}
Each query in the dynamic split is evaluated with web search enabled, allowing a maximum of two retrieval calls. As in the static split, the model is provided only with the query text and is not prompted or encouraged to invoke retrieval, as explained in the setup in Appendix \ref{app:eval-prompts}. Model outputs are evaluated via exact matching against the set of acceptable semantic answer variants. From an information retrieval perspective, this setting assesses whether models correctly detect information need and execute retrieval efficiently. In particular, we aim to analyse:
\begin{enumerate}
    \item \textit{Answer accuracy under necessary retrieval}, measuring end-to-end effectiveness.
    \item \textit{Retrieval recall and invocation rate}, capturing how often models correctly initiate web search when retrieval is required.
    \item \textit{Retrieval efficiency and cost-effectiveness}, measuring how many retrieval calls are needed to obtain correct, up-to-date information.
\end{enumerate}

\subsection{Experimental Setup}
We evaluate four LLMs equipped with native web search capabilities using their respective provider APIs, spanning both smaller and frontier model variants, under both static and dynamic retrieval regimes. Specifically, we experiment with OpenAI’s GPT-5-mini (gpt-5-mini-2025-08-07) and GPT-5 (gpt-5-2025-08-07). Similarly, we test Anthropic’s Claude Haiku 4.5 (claude-haiku-4-5-20251001) and Claude Sonnet 4.6 (claude-sonnet-4-6). For GPT, all requests specify only the browsing tool as \verb|tools=[{"type": "web_search"}]|, while for Claude, the tool is configured using \\ \verb|tools=[{"type": "web\_search\_20250305"}]|.

To ensure reproducibility and isolate retrieval behaviour, we set the temperature to 0 and allow a maximum of two web search calls per query, with search invocation determined autonomously by the model. This constraint reflects the retrieval assumptions of our benchmark: all queries in both splits are constructed such that correct answers are discoverable within the top one or two results of a standard web search. We interpret multiple search calls as successive retrieval attempts, analogous to query reformulation in classical IR. All other decoding and API parameters follow provider default settings. For each retrieval split, we use fixed, shared prompts that are issued verbatim to all models, followed by the task query. The exact prompt templates for static and dynamic settings are present in our repository for reproducibility \footnote{\url{https://anonymous.4open.science/r/look-it-up-0B20/prompts}}.

\definecolor{HalfBaked}{rgb}{0.498,0.701,0.835}
\definecolor{FuzzyWuzzyBrown}{rgb}{0.768,0.305,0.321}
\definecolor{SanMarino}{rgb}{0.298,0.447,0.69}
\definecolor{Shilo}{rgb}{0.901,0.647,0.658}

\begin{table}[t]
\centering
\caption{Accuracy by number of web search calls under the web-search enabled condition}
\label{tab:acc-web-calls}
\resizebox{\columnwidth}{!}{
\begin{tblr}{
  cells = {c},
  colspec = {p{0.28\columnwidth} p{0.24\columnwidth} p{0.24\columnwidth} p{0.24\columnwidth}},
  hline{1,2,7,12} = {-}{},
}
Model & 0 Calls & 1 Call & 2 Calls \\

\SetCell[c=4]{c}{\textbf{Static Split}} & & & \\
GPT-5-mini        & $0.943$ & $0.857$ & $0.652$ \\
Claude Haiku 4.5  & $0.786$ & $0.777$ & $0.351$ \\
GPT-5             & $0.924$ & $0.852$ & $0.712$ \\
Claude Sonnet 4.6 & $0.891$ & $0.843$ & $0.522$ \\

\SetCell[c=4]{c}{\textbf{Dynamic Split}} & & & \\
GPT-5-mini        & $0.000$ & $0.838$ & $0.684$ \\
Claude Haiku 4.5  & $0.000$ & $0.669$ & $0.125$ \\
GPT-5             & $0.000$ & $0.811$ & $0.692$ \\
Claude Sonnet 4.6 & $0.000$ & $0.761$ & $0.266$ \\
\end{tblr}
}
\end{table}

\section{Analysis and Result Discussion}
\subsection{How effective is internal web-based IR in improving LLM accuracy?} 
\label{sec:accuracy}
To evaluate the effectiveness of internal web search, we measure retrieval-enabled LLM performance in terms of downstream answer accuracy across both evaluation splits, as reported in Table~\ref{tab:acc}. Across models, enabling web search yields substantial accuracy gains, indicating that retrieval can meaningfully augment parametric knowledge. On the static split, all models show clear improvements when search is enabled. The relative gains are largest for smaller models, with GPT-5-mini achieving a \textit{62\% relative gain} and Claude Haiku 4.5 improving by approximately \textit{80\%}, while frontier models (GPT-5 and Claude Sonnet 4.6) show smaller but still consistent gains due to their strong baselines even without web search. From an IR perspective, these results indicate that internal web search can act as a verification mechanism, improving answer quality and accuracy.

Accuracy on the dynamic split remains lower overall. Although the split focuses on widely reported recent events, performance remains bounded across all models, with \textit{no model exceeding 72\% accuracy}. While frontier models (GPT-5 at 71.4\% and Claude Sonnet 4.6 at 66.2\%) outperform their smaller counterparts, these differences should be interpreted as end-to-end system differences, since each model family operates with its own provider-specific web-search infrastructure. The results therefore suggest that performance in this setting depends on both model capability and the associated retrieval system. This aligns with prior findings \cite{chen-etal-2025-acebench} that web search can improve performance, but does not guarantee precise or reliable retrieval.

As an external retrieval reference, we evaluate a custom RAG pipeline that decouples retrieval from the model’s internal decision-making \footnote{\url{https://github.com/openai/openai-cookbook/blob/main/examples/third_party/Web_search_with_google_api_bring_your_own_browser_tool.ipynb}}. This is intended to be interpreted as an end-to-end comparison between two different retrieval configurations rather than as a controlled comparison of search quality. For each query, we issue a web search using the Google Custom Search API, retrieve the top two results, and summarise their titles and snippets using the same LLM as in the internal web-search setting. The summary is then provided as context for answer generation, without enabling the model’s internal search tool.

As shown in Table~\ref{tab:external-baseline}, the external pipeline consistently outperforms internal web search on the dynamic split across all models by a substantial margin. The comparison provides an external reference point for the end-to-end effectiveness of internal web-based retrieval and indicates that explicitly controlled retrieval can substantially improve performance on these queries.

To further analyse retrieval effectiveness, we group results by the number of web search calls per query, as shown in Table~\ref{tab:acc-web-calls}. On the static split, accuracy is highest when no retrieval is used and generally degrades as additional calls are issued, indicating that unnecessary retrieval can harm performance. This effect is consistent across models, though less pronounced for frontier models. On the dynamic split, accuracy is zero across all models without retrieval, as expected. Performance peaks with a single web search call and declines with multiple calls. The decline is modest for GPT-5-mini and GPT-5 but substantially steeper for Claude models, particularly Claude Haiku 4.5, suggesting that repeated retrieval attempts are often associated with unresolved information need rather than additional performance gains. From an IR perspective, these results highlight the importance of early precision and selective retrieval, motivating the analysis that follows.

\subsection{Does web search effectively change factual outcomes when invoked?} 
Beyond aggregate accuracy, we analyse the utility and risk of retrieval by examining how invoking internal web search changes answer correctness on a per-query basis. Following the outcome-based framework of \cite{10.3389/frai.2024.1341697}, we classify each query according to whether retrieval improves, preserves, or degrades factual correctness. This analysis is restricted to static-split queries for which web search is invoked at least once, enabling direct comparison between retrieval-enabled and retrieval-disabled outcomes.

This analysis measures whether retrieval provides positive utility relative to a no-retrieval baseline, and whether it introduces errors when not strictly necessary. We categorise retrieval outcomes to analyse \textit{retrieval utility} and \textit{retrieval risk} as follows:

\begin{itemize}
\item \textit{Beneficial retrieval} includes cases where retrieval converts an incorrect answer into a correct one, as well as cases where retrieval preserves an already correct answer through verification, with low retrieval risk and high utility.
\item \textit{Harmful retrieval} includes cases where retrieval degrades a previously correct answer, as well as cases where the answer remains incorrect despite retrieval, indicating limited retrieval utility and high risk.
\end{itemize}

Tables~\ref{tab:gpt-ben-harm}--\ref{tab:claude-sonnet-ben-harm} report outcome distributions across all four models in a confusion-matrix-style format, including only those cases where web retrieval was triggered when available by each model.

Across models, retrieval utility consistently exceeds retrieval risk. This pattern holds for both smaller and frontier models, although their composition differs. Larger models exhibit a higher proportion of true beneficial retrieval, where retrieval converts incorrect answers into correct ones. In contrast, smaller models more frequently demonstrate verification behaviour, preserving correctness when retrieval is invoked despite an already correct answer. Across all models, retrieval-induced errors remain relatively rare, indicating that internal web search is generally safe to invoke.

Notably, the dominant form of retrieval risk is not retrieval-induced error, but failure to resolve information need. A substantial fraction of queries remain incorrect even after web search across all models. While this effect is a little less pronounced for frontier models, it remains a significant source of error. From an information retrieval perspective, these unresolved failures point to limitations in retrieval precision, query formulation, or evidence integration. Overall, while internal web search provides consistent utility, its effectiveness remains uneven.

\begin{table}[t]
\centering
\caption{Outcome transitions for GPT-5-mini on the static split with and without web search}
  \label{tab:gpt-ben-harm}
\resizebox{\columnwidth}{!}{
\begin{tblr}{
  cells = {c},
  colspec = {p{0.20\columnwidth} p{0.20\columnwidth} p{0.25\columnwidth} p{0.25\columnwidth}},
  cell{1}{3} = {c=2}{},
  cell{3}{1} = {r=2}{},
  cell{3}{3} = {HalfBaked,fg=white},
  cell{3}{4} = {FuzzyWuzzyBrown,fg=white},
  cell{4}{3} = {SanMarino,fg=white},
  cell{4}{4} = {Shilo,fg=white},
}
                         &           & \textit{with web search } &              \\
                         &           & Correct                   & Incorrect    \\
\textit{w/o web search } & Correct   & \textbf{218}              & \textbf{26}  \\
                         & Incorrect & \textbf{246}              & \textbf{117} 
\end{tblr}
}
\end{table}

\begin{table}[t]
\centering
\caption{Outcome transitions for Claude 4.5 Haiku on the static split with and without web search}
  \label{tab:claude-ben-harm}
\resizebox{\columnwidth}{!}{
\begin{tblr}{
  cells = {c},
  colspec = {p{0.20\columnwidth} p{0.20\columnwidth} p{0.25\columnwidth} p{0.25\columnwidth}},
  cell{1}{3} = {c=2}{},
  cell{3}{1} = {r=2}{},
  cell{3}{3} = {HalfBaked,fg=white},
  cell{3}{4} = {FuzzyWuzzyBrown,fg=white},
  cell{4}{3} = {SanMarino,fg=white},
  cell{4}{4} = {Shilo,fg=white},
}
                         &           & \textit{with web search } &              \\
                         &           & Correct                   & Incorrect    \\
\textit{w/o web search } & Correct   & \textbf{210}              & \textbf{33}  \\
                         & Incorrect & \textbf{295}              & \textbf{143} 
\end{tblr}
}
\end{table}

\begin{table}[t]
\centering
\caption{Outcome transitions for GPT-5 on the static split with and without web search}
  \label{tab:gpt5-ben-harm}
\resizebox{\columnwidth}{!}{
\begin{tblr}{
  cells = {c},
  colspec = {p{0.20\columnwidth} p{0.20\columnwidth} p{0.25\columnwidth} p{0.25\columnwidth}},
  cell{1}{3} = {c=2}{},
  cell{3}{1} = {r=2}{},
  cell{3}{3} = {HalfBaked,fg=white},
  cell{3}{4} = {FuzzyWuzzyBrown,fg=white},
  cell{4}{3} = {SanMarino,fg=white},
  cell{4}{4} = {Shilo,fg=white},
}
                         &           & \textit{with web search } &              \\
                         &           & Correct                   & Incorrect    \\
\textit{w/o web search } & Correct   & \textbf{215}              & \textbf{22}  \\
                         & Incorrect & \textbf{309}              & \textbf{109} 
\end{tblr}
}
\end{table}

\begin{table}[t]
\centering
\caption{Outcome transitions for Claude 4.6 Sonnet on the static split with and without web search}
  \label{tab:claude-sonnet-ben-harm}
\resizebox{\columnwidth}{!}{
\begin{tblr}{
  cells = {c},
  colspec = {p{0.20\columnwidth} p{0.20\columnwidth} p{0.25\columnwidth} p{0.25\columnwidth}},
  cell{1}{3} = {c=2}{},
  cell{3}{1} = {r=2}{},
  cell{3}{3} = {HalfBaked,fg=white},
  cell{3}{4} = {FuzzyWuzzyBrown,fg=white},
  cell{4}{3} = {SanMarino,fg=white},
  cell{4}{4} = {Shilo,fg=white},
}
                         &           & \textit{with web search } &              \\
                         &           & Correct                   & Incorrect    \\
\textit{w/o web search } & Correct   & \textbf{202}              & \textbf{28}  \\
                         & Incorrect & \textbf{262}              & \textbf{110} 
\end{tblr}
}
\end{table}

\begin{figure}[t]
\centering

\begin{subfigure}[b]{\columnwidth}
    \centering
    \includegraphics[width=0.8\linewidth]{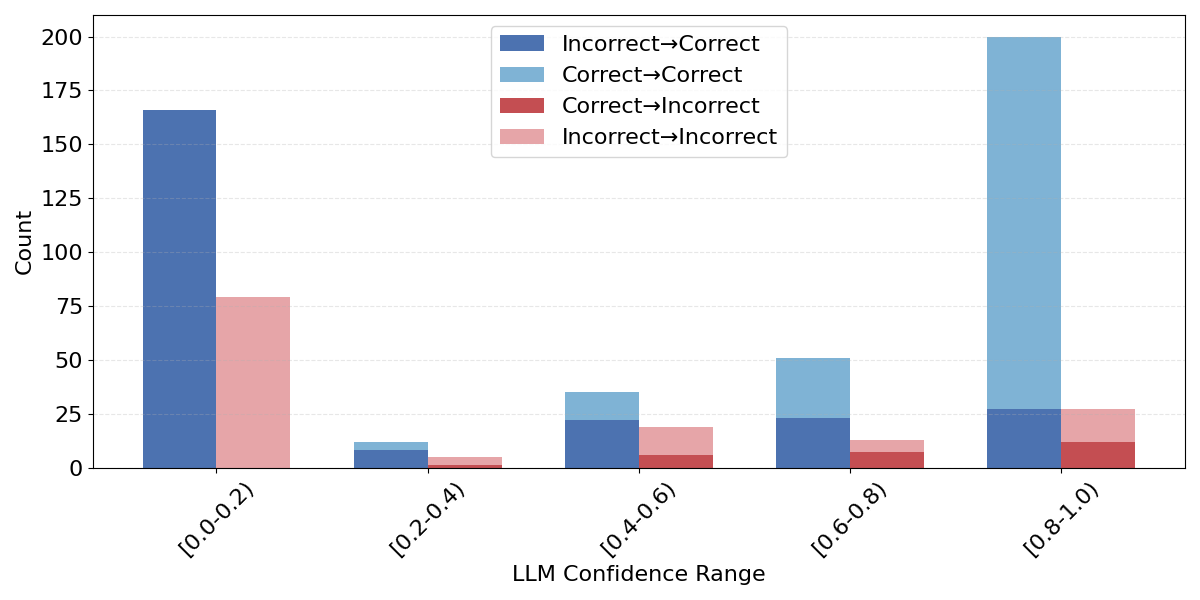}
    \Description{Outcome transitions binned by self-reported confidence for GPT-5-mini with and without web search}
    \caption{GPT-5-mini}
\end{subfigure}
\hfill
\begin{subfigure}[b]{\columnwidth}
    \centering
    \includegraphics[width=0.8\linewidth]{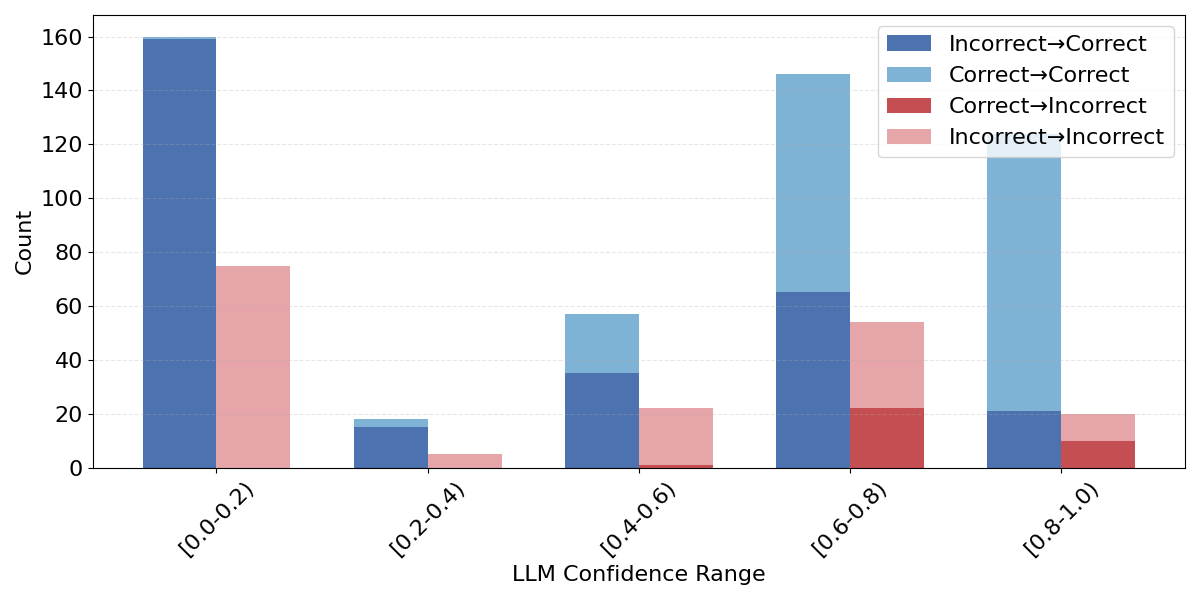}
    \Description{Outcome transitions binned by self-reported confidence for Claude Haiku 4.5 with and without web search}
    \caption{Claude Haiku 4.5}
\end{subfigure}

\vspace{0.5em}

\begin{subfigure}[b]{\columnwidth}
    \centering
    \includegraphics[width=0.8\linewidth]{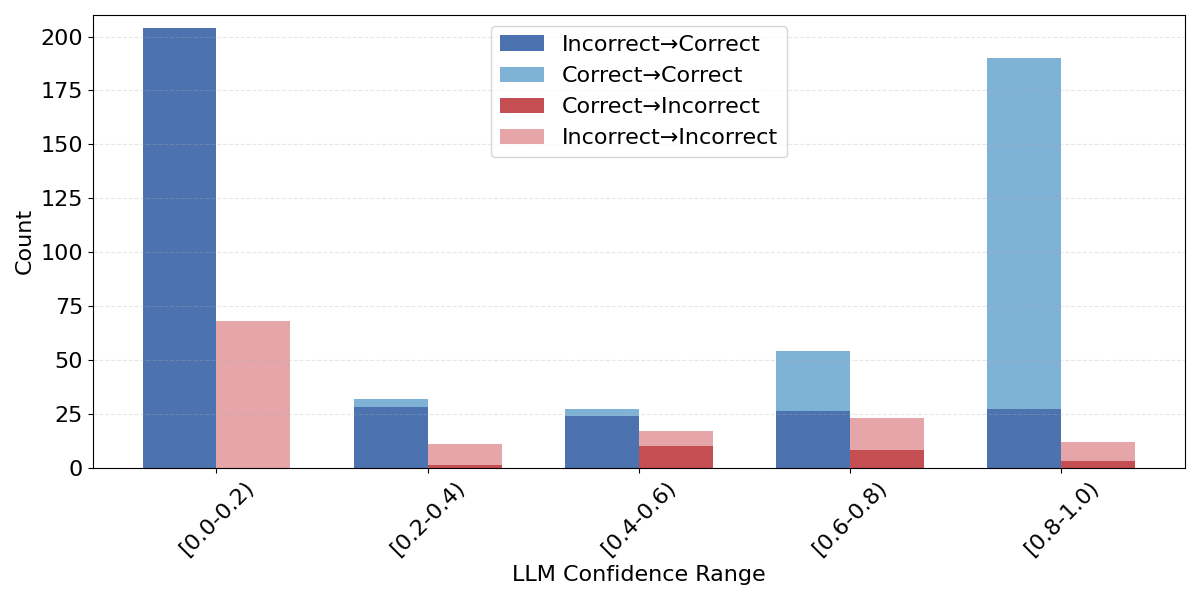}
    \Description{Outcome transitions binned by self-reported confidence for GPT-5 with and without web search}
    \caption{GPT-5}
\end{subfigure}
\hfill
\begin{subfigure}[b]{\columnwidth}
    \centering
    \includegraphics[width=0.8\linewidth]{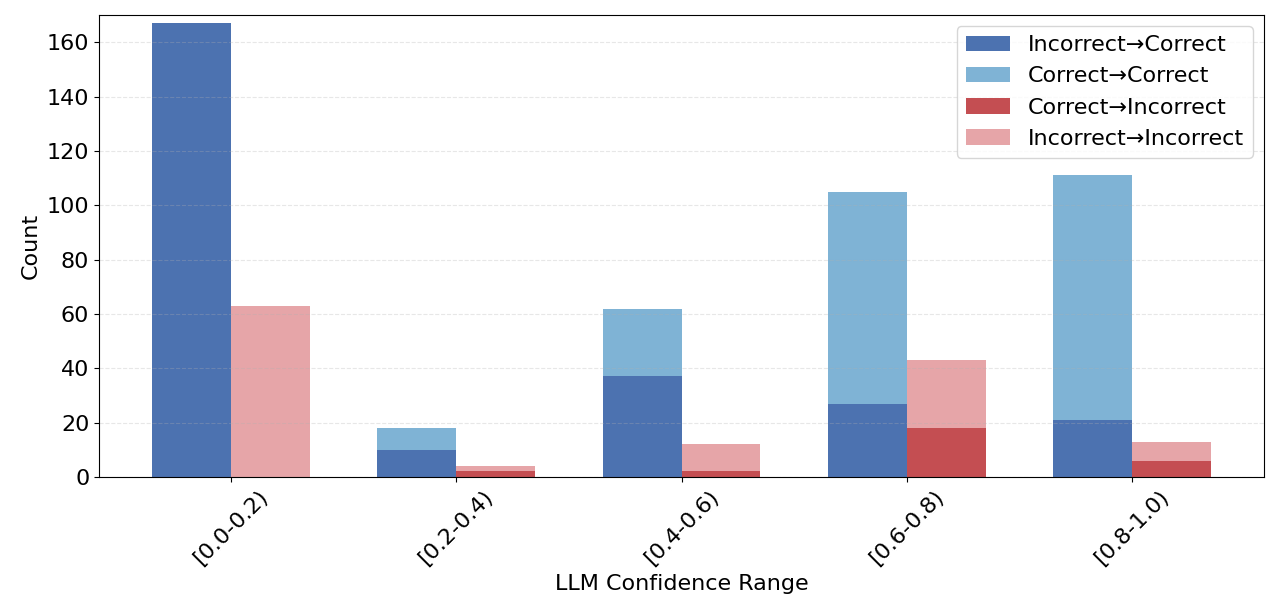}
    \Description{Outcome transitions binned by self-reported confidence for Claude Sonnet 4.6 with and without web search}
    \caption{Claude Sonnet 4.6}
\end{subfigure}

\caption{Outcome transitions across confidence bins}
\label{fig:conf-bins}
\end{figure}

\subsection{How well-calibrated are LLMs' verbalised confidence scores with and without web search?}

Because the evaluated systems are proprietary black-box models, we use verbalised confidence as a model-agnostic observable proxy for confidence, elicited on a standardized $[0,1]$ scale. This signal is not assumed to provide a direct measurement of internal uncertainty; rather, we evaluate whether the reported confidence is empirically aligned with correctness and whether it can serve as a useful retrieval-gating signal. To verify the stability of the verbalized scores, we also conducted a prompt-sensitivity analysis on 100 randomly selected queries using the two frontier models and found high consistency as described in Appendix \ref{app:conf}.

Table~\ref{tab:calibration} reports confidence calibration metrics on the static split, which we interpret as an evaluation of confidence as a retrieval gating signal. Without web search, all models exhibit reasonably well-calibrated confidence, reflected by low expected calibration error (ECE) and Brier scores, indicating alignment between stated confidence and empirical correctness. This suggests that confidence reliably reflects parametric knowledge adequacy and can support retrieval decisions.

When web search is enabled, overall accuracy improves, but confidence calibration deteriorates across models. All models become systematically overconfident, with substantial increases in calibration error and Brier scores. The effect is most pronounced for GPT-5-mini and Claude Haiku 4.5, while GPT-5 and Claude Sonnet 4.6 show comparatively smaller but still consistent degradation in ECE, alongside notable increases in Brier score. This indicates that access to retrieval alters the interpretation of confidence, decoupling self-reported certainty from actual retrieval effectiveness.

Overall, access to web search raises self-reported confidence without a corresponding improvement in calibration, weakening the reliability of verbalised confidence as a standalone signal for selective retrieval.

\begin{table}[t]
\centering
\caption{Calibration metrics for both models with and without web search on the static split}
\label{tab:calibration}

\setlength{\tabcolsep}{3pt}
\begin{tblr}{
  cells = {c},
  cell{2}{1} = {r=2}{},
  cell{2}{2} = {r},
  cell{3}{2} = {r},
  cell{4}{1} = {r=2}{},
  cell{4}{2} = {r},
  cell{5}{2} = {r},
  cell{6}{1} = {r=2}{},
  cell{8}{1} = {r=2}{},
  hline{1-2,4,6,8,10} = {-}{},
  column{1} = {l}, 
}
Model            &                          & ECE  & Brier \\
GPT-5-mini     & \textit{w/o web search}  & 0.10 & 0.11  \\
                 & \textit{with web search} & 0.27 & 0.31  \\
Claude Haiku 4.5 & \textit{w/o web search}  & 0.14 & 0.17  \\
                 & \textit{with web search} & 0.26 & 0.33  \\
GPT-5     & \textit{w/o web search}  & 0.10 & 0.15  \\
                 & \textit{with web search} & 0.17 & 0.29  \\
Claude Sonnet 4.6     & \textit{w/o web search}  & 0.08 & 0.16  \\
                 & \textit{with web search} & 0.21 & 0.41 
\end{tblr}

\end{table}

\subsection{Do models correctly invoke retrieval when information need is unavoidable?}
We evaluate model behaviour under retrieval necessity by analysing performance on the dynamic split, where external retrieval is required for correctness in all cases. Table~\ref{tab:dynamic-use} summarises retrieval invocation statistics, including the proportion of queries that trigger web search and the distribution of single versus multiple retrieval calls. All models invoke web search for the majority of queries, with invocation rates ranging from 87.5\% to 93.9\%. These rates indicate that models generally recognise strong temporal cues and detect information need. Frontier models achieve slightly higher invocation rates than smaller models. However, none of the models achieves perfect recall, and a non-trivial fraction of queries are still answered without retrieval despite requiring up-to-date information.

Retrieval depth reveals clear differences in strategy based on model families. GPT-5-mini and GPT-5 more frequently issue two retrieval calls, suggesting a more iterative approach resembling query reformulation. In contrast, Claude models, particularly Claude Haiku 4.5, resolve most queries with a single retrieval call, with second calls being relatively rare. As shown in Table~\ref{tab:acc-web-calls}, additional retrieval attempts do not consistently improve accuracy across models, indicating that multiple calls often reflect unresolved information need rather than effective refinement.

Although accuracy is reasonably high when retrieval is invoked, missed retrievals remain a clear failure mode. These false negatives occur despite explicit temporal cues and indicate incomplete retrieval recall. Overall, the results suggest that while models are generally effective at detecting retrieval necessity, improving retrieval triggering could yield direct gains in end-to-end effectiveness.

\begin{table}[t]
\centering
\caption{Retrieval statistics on the dynamic split, where retrieval is required for correctness}
  \label{tab:dynamic-use}
\resizebox{\columnwidth}{!}{
\begin{tblr}{
  cells = {c},
  colspec = {p{0.30\columnwidth} p{0.18\columnwidth} p{0.18\columnwidth} p{0.18\columnwidth} p{0.18\columnwidth}},
  cell{1}{1} = {r=2}{},
  cell{1}{2} = {r=2}{},
  cell{1}{3} = {r=2}{},
  cell{1}{4} = {c=2}{},
  hline{1,3,7} = {-}{},
  hline{2} = {4-5}{},
}
Model            & No. of dynamic queries & Retrieval invocation Rate & Retrieval depth (no. of calls) &     \\
                 &                        &                 & One                           & Two \\
GPT-5-mini     & 288                    & 87.5\%    & 154                           & 98  \\
Claude Haiku 4.5 & 288                    & 91.0\%    & 254                           & 8   \\
GPT-5 & 288                    & 92.1\%    & 178                           & 87   \\
Claude Sonnet 4.6 & 288                    & 93.9\%    & 214                           & 56   
\end{tblr}
}
\end{table}

\begin{table}[t]
\centering
\caption{Distribution of retrieval failure types across models and splits}
\label{tab:failure-types}
\resizebox{\columnwidth}{!}{
\begin{tblr}{
  cells = {c},
  colspec = {p{0.22\columnwidth} p{0.26\columnwidth} p{0.17\columnwidth} p{0.17\columnwidth} p{0.18\columnwidth}},
  cell{2}{1} = {r=2}{},
  cell{4}{1} = {r=2}{},
  cell{6}{1} = {r=2}{},
  cell{8}{1} = {r=2}{},
  hline{1,2,4,6,8,10} = {-}{},
}
Model            & Split   & QF (\%) & SF (\%) & EF (\%) \\
GPT-5-mini       & Static  & 51.4    & 33.6    & 15.0    \\
                 & Dynamic & 44.6    & 44.6    & 10.8    \\
Claude Haiku 4.5 & Static  & 57.8    & 27.5    & 14.7    \\
                 & Dynamic & 39.4    & 47.9    & 12.7    \\
GPT-5 & Static  & 49.8    & 39.5    & 10.7    \\
                 & Dynamic & 46.6    & 53.4    & 6.8    \\
Claude Sonnet 4.6 & Static  & 59.9    & 23.0    & 17.1    \\
                 & Dynamic & 43.7    & 40.1    & 16.2    \\
\end{tblr}
}
\end{table}

\subsection{Where and why does retrieval fail?}

To understand the limits of internal web search beyond aggregate accuracy and missed retrieval, we conduct a targeted error analysis over failed retrieval cases where web search was invoked but the final answer was incorrect. We manually annotate cases into three mutually exclusive failure categories corresponding to core stages of the retrieval pipeline, as shown in Table~\ref{tab:failure-types}. Examples of each failure type are given in Table~\ref{tab:error_examples}. Two independent annotators with graduate-level expertise applied predefined annotation criteria on a random sample of 150 incorrect examples per model for the static split and 75 for the dynamic split achieving Cohen's $\kappa=0.79$, with disagreements resolved through discussion and consensus. The annotation protocol and inter-annotator agreement are reported in Appendix~\ref{app:failure-annotation}.

\begin{itemize}
    \item \textit{Query formulation failures (QF):} Occur when the issued search query underspecifies the information need, omits critical entities or constraints, or reformulates the question in a way that is not relevant.
    \item \textit{Source quality failures (SF):} Capture cases where the query is well formed, but retrieved sources are outdated, low-quality, or insufficient to support the correct answer. 
    \item \textit{Evidence integration failures (EF):} Cases where the query is correctly formulated and relevant information is retrieved, but is not correctly reflected in the final response.
\end{itemize}

Across models, query formulation failures constitute the largest source of error on the static split, accounting for roughly half of all failures. In the dynamic split, however, source quality failures increase substantially and become comparable to or exceed query formulation failures, particularly for frontier models. Evidence integration failures are consistently the least frequent, indicating that once relevant and reliable information is retrieved, models are generally capable of incorporating it correctly.

Overall, these results suggest that the primary bottlenecks in internal web search lie in query formulation and source selection rather than answer synthesis. This suggests that improving internal web search may benefit from stronger query formulation and better temporal- and credibility-aware source selection.

\begin{table*}[t]
\centering
\caption{Cost-accuracy metrics for web search usage by models}
  \label{tab:cost}
\resizebox{\textwidth}{!}{
\begin{tblr}{
  cells = {c},
  colspec = {
    p{0.06\textwidth}  
    p{0.15\textwidth}  
    p{0.08\textwidth}  
    p{0.10\textwidth}  
    p{0.10\textwidth}  
    p{0.10\textwidth}  
    p{0.10\textwidth}  
    p{0.12\textwidth}  
    p{0.12\textwidth}  
    p{0.12\textwidth}  
  },
  cell{2}{1} = {r=4}{},
  cell{6}{1} = {r=4}{},
  hline{1} = {1-10}{},
  hline{2,6,10} = {-}{}
}
Split   & Model            & Total Queries & Total Web Search Calls & Total Extra Cost (USD) & Overall Accuracy (\%) & Average Calls/Query & Average Extra Cost per Query (USD) & Average Cost per Strict Improvement (USD) & Average Cost per Inclusive Improvement (USD) \\
Static  & GPT 5 mini     & 783           & 844                    & 8.44                   & 84.6                  & 1.077               & 0.011                              & 0.034                                     & 0.013                                        \\
        & Claude Haiku 4.5 & 783           & 738                    & 7.38                   & 74.7                  & 0.942               & 0.009                              & 0.022                                     & 0.012                                        \\
        & GPT-5 & 783           & 801                    & 8.01                   & 90.7                  & 1.023               & 0.010                              & 0.031                                     & 0.013                                        \\
        & Claude Sonnet 4.6 & 783           & 774                    & 7.74                   & 86.8                  & 0.989               & 0.010                              & 0.026                                     & 0.012                                        \\
Dynamic & GPT 5 mini     & 288           & 350                    & 3.50                   & 68.4                  & 1.216               & 0.012                              & 0.021                                     & -                                            \\
        & Claude Haiku 4.5 & 288           & 270                    & 2.70                   & 60.8                  & 0.938               & 0.009                              & 0.014                                     & -               \\
    & GPT-5 & 288           & 439                    & 4.39                   & 71.4                  & 1.524               & 0.015                              & 0.032                                     & -     \\
    & Claude Sonnet 4.6 & 288           & 283                    & 2.83                   & 66.2                  & 0.982               & 0.010                              & 0.017                                     & -     
\end{tblr}
}
\end{table*}

\subsection{How cost-effective is web search in LLMs?}
Table~\ref{tab:cost} summarises web search usage and cost efficiency of retrieval across both dataset splits. Total retrieval cost is computed as the number of web calls multiplied by the per-call price (\$10 per 1k calls for both OpenAI and Anthropic API tools). We report only the additional cost from web search invocation and do not account for token-level differences, since queries and responses are short.

To better reflect selective retrieval behaviour, we additionally report cost-normalised improvement metrics. \textit{Strict improvement cost} measures the average cost required to convert an incorrect answer into a correct one relative to the no-search baseline. \textit{Inclusive improvement cost} further includes cases where retrieval preserves an already correct answer through verification. For the dynamic split, inclusive improvement does not apply.

Two observations stand out. First, web-based retrieval costs are modest in absolute terms, with average per-query cost around one cent across all models and splits, despite roughly one search call per query. Second, cost-per-improvement varies more sharply across models. Claude models consistently achieve lower cost per corrected answer than GPT models, reflecting more selective retrieval usage, while GPT models incur higher costs due to more frequent or repeated calls. This pattern persists even for frontier models, where higher accuracy does not translate into better cost-efficiency.

On the dynamic split, strict improvement costs decrease across all models, indicating that retrieval is more cost-aligned when information need is explicit. However, differences in efficiency remain, with Claude models retaining a consistent advantage. Overall, while web search is inexpensive, its cost-effectiveness depends strongly on retrieval selectivity and efficiency rather than model scale alone. 

\section{Practical Implications and Guidance}
Our findings yield several practical implications for deploying LLMs with web-based retrieval capabilities.

\subsection{Cost–accuracy trade-off}
Our results show that closed-source LLMs with web-based retrieval can substantially improve factual accuracy at low marginal cost, but that its effectiveness depends critically on information need and retrieval depth. For queries with well-defined, stable information needs, a single, high-precision retrieval step offers a strong accuracy–cost trade-off. In such cases, enabling retrieval with a strict upper bound of one call functions effectively as a verification layer. In contrast, for queries involving rapidly evolving information, repeated retrieval attempts exhibit diminishing returns across models. Multiple retrieval calls often reflect unresolved information need or ineffective query formulation rather than improved evidence discovery. Deeper retrieval without improved precision increases cost while offering limited utility. We recommend that deployments that expose models to live data streams on the web should instead use external tools or scaffolding to feed information into LLMs, rather than relying on inbuilt search capabilities.

\subsection{Inbuilt web search vs external retrieval}
Our analysis shows the difference between model-initiated web search and externally orchestrated IR pipelines. Internal web search is most suitable for lightweight, low-latency applications such as conversational assistants or interactive querying where simplicity and responsiveness are prioritised. Externally scaffolded retrieval pipelines, including classical RAG systems and application-level query planners, offer greater control over query formulation, ranking, and source selection. Even a simple external pipeline outperforms internal web search across all evaluated models, as shown in Section~\ref{sec:accuracy}. External systems are better suited for settings where retrieval must be auditable, domain-restricted, or grounded in authoritative corpora, such as scientific, medical, or legal applications. Internal web search prioritises agility and opportunistic retrieval, whereas external pipelines prioritise governance, reproducibility, and source quality. The appropriate choice depends on whether the task values retrieval speed or retrieval control.

\subsection{Selective invocation strategies}
Selective retrieval as performed by LLMs emerges as an effective compromise between always-retrieve and no-retrieve strategies. Our results show that confidence-gated retrieval achieves favourable accuracy–cost trade-offs, indicating that internal factual confidence provides a useful, albeit imperfect, signal of information need. However, confidence calibration deteriorates once retrieval is enabled across models, weakening its reliability as a stand-alone signal. Consequently, production systems should avoid relying solely on verbalised confidence when retrieval is available. Instead, confidence should be combined with additional signals such as retrieval redundancy, source agreement, or citation quality to support robust retrieval decisions and downstream verification.

\subsection{Provider-specific behaviours}

The evaluated systems exhibit distinct retrieval behaviours with practical implications. Smaller models tend to rely more on retrieval for correction, while frontier models more often use retrieval for verification given stronger parametric knowledge. Across the evaluated deployments, Claude systems issue fewer retrieval calls and achieve lower cost per corrected answer, while GPT systems invoke retrieval more frequently and perform more iterative calls. These differences suggest that retrieval behaviour varies across model--search-system combinations and may reflect both model-level retrieval policies and differences in the underlying provider infrastructure. From a deployment perspective, Claude-based configurations may therefore be attractive when predictable retrieval cost is prioritised, whereas more frequent retrieval by GPT-based configurations may be beneficial in settings where additional evidence gathering is preferred. 

\section{Conclusion}
Overall, our paper introduces a principled, externally observable framework for evaluating how closed-source LLMs decide when to invoke internal web search and how effectively they use retrieved evidence to produce correct answers. By combining static and dynamic factual splits with outcome transition analysis and confidence-based selective routing, we isolate three key components of web-enabled LLM behaviour: internal uncertainty estimation, retrieval invocation decisions, and downstream factual reliability. Crucially, this evaluation does not require access to model internals, tool prompts, or provider-specific configuration, making it broadly applicable across both smaller and frontier systems.

Our results show that internal web search yields substantial accuracy gains across models, but also introduces systematic calibration failures. While access to retrieval improves factual correctness, it inflates self-reported confidence even when retrieval does not resolve uncertainty. From an information retrieval perspective, this reveals a misalignment between retrieval effectiveness and confidence signalling, with implications for retrieval gating, verification, and downstream decision-making.

More broadly, our findings highlight the interaction between parametric knowledge, uncertainty estimation, and external retrieval in web-enabled LLMs. We show that retrieval effectiveness is primarily constrained by query formulation and source selection rather than model scale alone, and that improvements in retrieval triggering and efficiency offer direct gains in end-to-end performance. To support continued evaluation under evolving real-world conditions, we additionally provide a lightweight mechanism to periodically refresh the dynamic split, ensuring sustained alignment with current information. We release our full dataset and evaluation code, providing a transparent benchmark for measuring alignment between information need and retrieval behaviour in closed-source LLMs.

\section{Limitations and Future Work}
While our benchmark provides a structured evaluation of internal web search in LLMs, certain limitations remain that point to directions for future research.

\begin{itemize}
\item \textit{Experimental coverage:} Our evaluation focuses on studying native, tool-integrated web search in LLMs. The methodology can be extended to other commercial systems and to open-source models with access to internal tool parameters.

\item \textit{Limitations of live-web evaluation.} Because the evaluated systems access provider-controlled live search infrastructure, the exact retrieval trajectory cannot be guaranteed to remain identical across time. The static split is designed to remain temporally stable, whereas the dynamic split is explicitly versioned by snapshot date and can be refreshed in future releases.

\item \textit{Retrieval complexity and confidence elicitation:} We limit retrieval depth to at most two calls and do not evaluate multi-hop, compositional, or adversarial queries requiring iterative reformulation or reasoning over multiple sources. Also, we use only verbalized confidence signals for our analysis, which can be expanded to logits or other signals. 

\item \textit{Language scope:} Our benchmark is restricted to English-language queries, sources, and responses. Consequently, the observed retrieval behaviours, failure modes, and calibration patterns may not generalise to multilingual or cross-lingual settings.
\end{itemize}

\bibliographystyle{ACM-Reference-Format}
\bibliography{sample-base}

@inproceedings{xu-etal-2025-alignment,
    title = "Alignment for Efficient Tool Calling of Large Language Models",
    author = "Xu, Hongshen  and
      Wang, Zihan  and
      Zhu, Zichen  and
      Pan, Lei  and
      Chen, Xingyu  and
      Fan, Shuai  and
      Chen, Lu  and
      Yu, Kai",
    editor = "Christodoulopoulos, Christos  and
      Chakraborty, Tanmoy  and
      Rose, Carolyn  and
      Peng, Violet",
    booktitle = "Proceedings of the 2025 Conference on Empirical Methods in Natural Language Processing",
    month = nov,
    year = "2025",
    address = "Suzhou, China",
    publisher = "Association for Computational Linguistics",
    url = "https://aclanthology.org/2025.emnlp-main.898/",
    doi = "10.18653/v1/2025.emnlp-main.898",
    pages = "17787--17803",
    ISBN = "979-8-89176-332-6"
}

@misc{zheng2025deepresearcherscalingdeepresearch,
      title={DeepResearcher: Scaling Deep Research via Reinforcement Learning in Real-world Environments}, 
      author={Yuxiang Zheng and Dayuan Fu and Xiangkun Hu and Xiaojie Cai and Lyumanshan Ye and Pengrui Lu and Pengfei Liu},
      year={2025},
      eprint={2504.03160},
      archivePrefix={arXiv},
      primaryClass={cs.AI},
      url={https://arxiv.org/abs/2504.03160}, 
}

@misc{cheng2025dailyqabenchmarkevaluateweb,
      title={DailyQA: A Benchmark to Evaluate Web Retrieval Augmented LLMs Based on Capturing Real-World Changes}, 
      author={Jiehan Cheng and Zhicheng Dou},
      year={2025},
      eprint={2505.17162},
      archivePrefix={arXiv},
      primaryClass={cs.IR},
      url={https://arxiv.org/abs/2505.17162}, 
}

@inproceedings{kim-etal-2024-rada,
    title = "{R}a{DA}: Retrieval-augmented Web Agent Planning with {LLM}s",
    author = "Kim, Minsoo  and
      Bursztyn, Victor  and
      Koh, Eunyee  and
      Guo, Shunan  and
      Hwang, Seung-won",
    editor = "Ku, Lun-Wei  and
      Martins, Andre  and
      Srikumar, Vivek",
    booktitle = "Findings of the Association for Computational Linguistics: ACL 2024",
    month = aug,
    year = "2024",
    address = "Bangkok, Thailand",
    publisher = "Association for Computational Linguistics",
    url = "https://aclanthology.org/2024.findings-acl.802/",
    doi = "10.18653/v1/2024.findings-acl.802",
    pages = "13511--13525"
}

@misc{kale2025miragemasterymemorizationtricks,
      title={Mirage of Mastery: Memorization Tricks LLMs into Artificially Inflated Self-Knowledge}, 
      author={Sahil Kale and Vijaykant Nadadur},
      year={2025},
      eprint={2506.18998},
      archivePrefix={arXiv},
      primaryClass={cs.CL},
      url={https://arxiv.org/abs/2506.18998}, 
}

@misc{dechezelles2025browsergymecosystemwebagent,
      title={The BrowserGym Ecosystem for Web Agent Research}, 
      author={Thibault Le Sellier De Chezelles and Maxime Gasse and Alexandre Drouin and Massimo Caccia and Léo Boisvert and Megh Thakkar and Tom Marty and Rim Assouel and Sahar Omidi Shayegan and Lawrence Keunho Jang and Xing Han Lù and Ori Yoran and Dehan Kong and Frank F. Xu and Siva Reddy and Quentin Cappart and Graham Neubig and Ruslan Salakhutdinov and Nicolas Chapados and Alexandre Lacoste},
      year={2025},
      eprint={2412.05467},
      archivePrefix={arXiv},
      primaryClass={cs.LG},
      url={https://arxiv.org/abs/2412.05467}, 
}

@misc{miroyan2025searcharenaanalyzingsearchaugmented,
      title={Search Arena: Analyzing Search-Augmented LLMs}, 
      author={Mihran Miroyan and Tsung-Han Wu and Logan King and Tianle Li and Jiayi Pan and Xinyan Hu and Wei-Lin Chiang and Anastasios N. Angelopoulos and Trevor Darrell and Narges Norouzi and Joseph E. Gonzalez},
      year={2025},
      eprint={2506.05334},
      archivePrefix={arXiv},
      primaryClass={cs.CL},
      url={https://arxiv.org/abs/2506.05334}, 
}

@misc{wei2025browsecompsimplechallengingbenchmark,
      title={BrowseComp: A Simple Yet Challenging Benchmark for Browsing Agents}, 
      author={Jason Wei and Zhiqing Sun and Spencer Papay and Scott McKinney and Jeffrey Han and Isa Fulford and Hyung Won Chung and Alex Tachard Passos and William Fedus and Amelia Glaese},
      year={2025},
      eprint={2504.12516},
      archivePrefix={arXiv},
      primaryClass={cs.CL},
      url={https://arxiv.org/abs/2504.12516}, 
}

@misc{zhou2025browsecompzhbenchmarkingwebbrowsing,
      title={BrowseComp-ZH: Benchmarking Web Browsing Ability of Large Language Models in Chinese}, 
      author={Peilin Zhou and Bruce Leon and Xiang Ying and Can Zhang and Yifan Shao and Qichen Ye and Dading Chong and Zhiling Jin and Chenxuan Xie and Meng Cao and Yuxin Gu and Sixin Hong and Jing Ren and Jian Chen and Chao Liu and Yining Hua},
      year={2025},
      eprint={2504.19314},
      archivePrefix={arXiv},
      primaryClass={cs.CL},
      url={https://arxiv.org/abs/2504.19314}, 
}

@inproceedings{mind2web,
 author = {Deng, Xiang and Gu, Yu and Zheng, Boyuan and Chen, Shijie and Stevens, Sam and Wang, Boshi and Sun, Huan and Su, Yu},
 booktitle = {Advances in Neural Information Processing Systems},
 editor = {A. Oh and T. Naumann and A. Globerson and K. Saenko and M. Hardt and S. Levine},
 pages = {28091--28114},
 publisher = {Curran Associates, Inc.},
 title = {Mind2Web: Towards a Generalist Agent for the Web},
 url = {https://proceedings.neurips.cc/paper_files/paper/2023/file/5950bf290a1570ea401bf98882128160-Paper-Datasets_and_Benchmarks.pdf},
 volume = {36},
 year = {2023}
}

@misc{gou2025mind2web2evaluatingagentic,
      title={Mind2Web 2: Evaluating Agentic Search with Agent-as-a-Judge}, 
      author={Boyu Gou and Zanming Huang and Yuting Ning and Yu Gu and Michael Lin and Weijian Qi and Andrei Kopanev and Botao Yu and Bernal Jiménez Gutiérrez and Yiheng Shu and Chan Hee Song and Jiaman Wu and Shijie Chen and Hanane Nour Moussa and Tianshu Zhang and Jian Xie and Yifei Li and Tianci Xue and Zeyi Liao and Kai Zhang and Boyuan Zheng and Zhaowei Cai and Viktor Rozgic and Morteza Ziyadi and Huan Sun and Yu Su},
      year={2025},
      eprint={2506.21506},
      archivePrefix={arXiv},
      primaryClass={cs.AI},
      url={https://arxiv.org/abs/2506.21506}, 
}

@inproceedings{wu-etal-2025-webwalker,
    title = "{W}eb{W}alker: Benchmarking {LLM}s in Web Traversal",
    author = "Wu, Jialong  and
      Yin, Wenbiao  and
      Jiang, Yong  and
      Wang, Zhenglin  and
      Xi, Zekun  and
      Fang, Runnan  and
      Zhang, Linhai  and
      He, Yulan  and
      Zhou, Deyu  and
      Xie, Pengjun  and
      Huang, Fei",
    editor = "Che, Wanxiang  and
      Nabende, Joyce  and
      Shutova, Ekaterina  and
      Pilehvar, Mohammad Taher",
    booktitle = "Proceedings of the 63rd Annual Meeting of the Association for Computational Linguistics (Volume 1: Long Papers)",
    month = jul,
    year = "2025",
    address = "Vienna, Austria",
    publisher = "Association for Computational Linguistics",
    url = "https://aclanthology.org/2025.acl-long.508/",
    doi = "10.18653/v1/2025.acl-long.508",
    pages = "10290--10305",
    ISBN = "979-8-89176-251-0"
}

@inproceedings{t-eval-chen-etal-2024-eval,
    title = "{T}-Eval: Evaluating the Tool Utilization Capability of Large Language Models Step by Step",
    author = "Chen, Zehui  and
      Du, Weihua  and
      Zhang, Wenwei  and
      Liu, Kuikun  and
      Liu, Jiangning  and
      Zheng, Miao  and
      Zhuo, Jingming  and
      Zhang, Songyang  and
      Lin, Dahua  and
      Chen, Kai  and
      Zhao, Feng",
    editor = "Ku, Lun-Wei  and
      Martins, Andre  and
      Srikumar, Vivek",
    booktitle = "Proceedings of the 62nd Annual Meeting of the Association for Computational Linguistics (Volume 1: Long Papers)",
    month = aug,
    year = "2024",
    address = "Bangkok, Thailand",
    publisher = "Association for Computational Linguistics",
    url = "https://aclanthology.org/2024.acl-long.515/",
    doi = "10.18653/v1/2024.acl-long.515",
    pages = "9510--9529"
}

@inproceedings{guo-etal-2024-stabletoolbench,
    title = "{S}table{T}ool{B}ench: Towards Stable Large-Scale Benchmarking on Tool Learning of Large Language Models",
    author = "Guo, Zhicheng  and
      Cheng, Sijie  and
      Wang, Hao  and
      Liang, Shihao  and
      Qin, Yujia  and
      Li, Peng  and
      Liu, Zhiyuan  and
      Sun, Maosong  and
      Liu, Yang",
    editor = "Ku, Lun-Wei  and
      Martins, Andre  and
      Srikumar, Vivek",
    booktitle = "Findings of the Association for Computational Linguistics: ACL 2024",
    month = aug,
    year = "2024",
    address = "Bangkok, Thailand",
    publisher = "Association for Computational Linguistics",
    url = "https://aclanthology.org/2024.findings-acl.664/",
    doi = "10.18653/v1/2024.findings-acl.664",
    pages = "11143--11156"
}

@inproceedings{chen-etal-2025-acebench,
    title = "{ACEB}ench: A Comprehensive Evaluation of {LLM} Tool Usage",
    author = "Chen, Chen  and
      Hao, Xinlong  and
      Liu, Weiwen  and
      Huang, Xu  and
      Zeng, Xingshan  and
      Yu, Shuai  and
      Li, Dexun  and
      Huang, Yuefeng  and
      Liu, Xiangcheng  and
      Xinzhi, Wang  and
      Liu, Wu",
    editor = "Christodoulopoulos, Christos  and
      Chakraborty, Tanmoy  and
      Rose, Carolyn  and
      Peng, Violet",
    booktitle = "Findings of the Association for Computational Linguistics: EMNLP 2025",
    month = nov,
    year = "2025",
    address = "Suzhou, China",
    publisher = "Association for Computational Linguistics",
    url = "https://aclanthology.org/2025.findings-emnlp.697/",
    doi = "10.18653/v1/2025.findings-emnlp.697",
    pages = "12970--12998",
    ISBN = "979-8-89176-335-7"
}

@misc{zhang2025agentsafetybenchevaluatingsafetyllm,
      title={Agent-SafetyBench: Evaluating the Safety of LLM Agents}, 
      author={Zhexin Zhang and Shiyao Cui and Yida Lu and Jingzhuo Zhou and Junxiao Yang and Hongning Wang and Minlie Huang},
      year={2025},
      eprint={2412.14470},
      archivePrefix={arXiv},
      primaryClass={cs.CL},
      url={https://arxiv.org/abs/2412.14470}, 
}

@misc{andriushchenko2025agentharmbenchmarkmeasuringharmfulness,
      title={AgentHarm: A Benchmark for Measuring Harmfulness of LLM Agents}, 
      author={Maksym Andriushchenko and Alexandra Souly and Mateusz Dziemian and Derek Duenas and Maxwell Lin and Justin Wang and Dan Hendrycks and Andy Zou and Zico Kolter and Matt Fredrikson and Eric Winsor and Jerome Wynne and Yarin Gal and Xander Davies},
      year={2025},
      eprint={2410.09024},
      archivePrefix={arXiv},
      primaryClass={cs.LG},
      url={https://arxiv.org/abs/2410.09024}, 
}

@inproceedings{ke-etal-2025-adaptation,
    title = "Adaptation of Large Language Models",
    author = "Ke, Zixuan  and
      Ming, Yifei  and
      Joty, Shafiq",
    editor = "Lomeli, Maria  and
      Swayamdipta, Swabha  and
      Zhang, Rui",
    booktitle = "Proceedings of the 2025 Annual Conference of the Nations of the Americas Chapter of the Association for Computational Linguistics: Human Language Technologies (Volume 5: Tutorial Abstracts)",
    month = may,
    year = "2025",
    address = "Albuquerque, New Mexico",
    publisher = "Association for Computational Linguistics",
    url = "https://aclanthology.org/2025.naacl-tutorial.5/",
    doi = "10.18653/v1/2025.naacl-tutorial.5",
    pages = "30--37",
    ISBN = "979-8-89176-193-3"
}

@inproceedings{li-etal-2024-fundamental,
    title = "Fundamental Capabilities of Large Language Models and their Applications in Domain Scenarios: A Survey",
    author = "Li, Jiawei  and
      Yang, Yizhe  and
      Bai, Yu  and
      Zhou, Xiaofeng  and
      Li, Yinghao  and
      Sun, Huashan  and
      Liu, Yuhang  and
      Si, Xingpeng  and
      Ye, Yuhao  and
      Wu, Yixiao  and
      Lin, Yiguan  and
      Xu, Bin  and
      Ren, Bowen  and
      Feng, Chong  and
      Gao, Yang  and
      Huang, Heyan",
    editor = "Ku, Lun-Wei  and
      Martins, Andre  and
      Srikumar, Vivek",
    booktitle = "Proceedings of the 62nd Annual Meeting of the Association for Computational Linguistics (Volume 1: Long Papers)",
    month = aug,
    year = "2024",
    address = "Bangkok, Thailand",
    publisher = "Association for Computational Linguistics",
    url = "https://aclanthology.org/2024.acl-long.599/",
    doi = "10.18653/v1/2024.acl-long.599",
    pages = "11116--11141"
}

@article{kale2024faqgen,
  author       = {Kale, Sahil and Khaire, Gautam and Patankar, Jay},
  title        = {FAQ‑Gen: An automated system to generate domain‑specific FAQs to aid content comprehension},
  journal      = {Journal of Computer‑Assisted Linguistic Research},
  volume       = {8},
  number       = {1},
  pages        = {23--50},
  year         = {2024},
  doi          = {10.4995/jclr.2024.21178},
  url          = {https://doi.org/10.4995/jclr.2024.21178}
}

@inproceedings{yeginbergen-etal-2025-dynamic,
    title = "Dynamic Knowledge Integration for Evidence-Driven Counter-Argument Generation with Large Language Models",
    author = "Yeginbergen, Anar  and
      Oronoz, Maite  and
      Agerri, Rodrigo",
    editor = "Che, Wanxiang  and
      Nabende, Joyce  and
      Shutova, Ekaterina  and
      Pilehvar, Mohammad Taher",
    booktitle = "Findings of the Association for Computational Linguistics: ACL 2025",
    month = jul,
    year = "2025",
    address = "Vienna, Austria",
    publisher = "Association for Computational Linguistics",
    url = "https://aclanthology.org/2025.findings-acl.1161/",
    doi = "10.18653/v1/2025.findings-acl.1161",
    pages = "22568--22584",
    ISBN = "979-8-89176-256-5"
}

@inproceedings{wang-etal-2025-bring,
    title = "Bring Your Own Knowledge: A Survey of Methods for {LLM} Knowledge Expansion",
    author = {Wang, Mingyang  and
      Stoll, Alisa  and
      Lange, Lukas  and
      Adel, Heike  and
      Schuetze, Hinrich  and
      Str{\"o}tgen, Jannik},
    editor = "Jia, Robin  and
      Wallace, Eric  and
      Huang, Yangsibo  and
      Pimentel, Tiago  and
      Maini, Pratyush  and
      Dankers, Verna  and
      Wei, Johnny  and
      Lesci, Pietro",
    booktitle = "Proceedings of the First Workshop on Large Language Model Memorization (L2M2)",
    month = aug,
    year = "2025",
    address = "Vienna, Austria",
    publisher = "Association for Computational Linguistics",
    url = "https://aclanthology.org/2025.l2m2-1.12/",
    doi = "10.18653/v1/2025.l2m2-1.12",
    pages = "150--168",
    ISBN = "979-8-89176-278-7"
}

@inproceedings{ko-etal-2024-growover,
    title = "{G}row{OVER}: How Can {LLM}s Adapt to Growing Real-World Knowledge?",
    author = "Ko, Dayoon  and
      Kim, Jinyoung  and
      Choi, Hahyeon  and
      Kim, Gunhee",
    editor = "Ku, Lun-Wei  and
      Martins, Andre  and
      Srikumar, Vivek",
    booktitle = "Proceedings of the 62nd Annual Meeting of the Association for Computational Linguistics (Volume 1: Long Papers)",
    month = aug,
    year = "2024",
    address = "Bangkok, Thailand",
    publisher = "Association for Computational Linguistics",
    url = "https://aclanthology.org/2024.acl-long.181/",
    doi = "10.18653/v1/2024.acl-long.181",
    pages = "3282--3308"
}

@inproceedings{liu-etal-2025-real,
    title = "Real-time Ad Retrieval via {LLM}-generative Commercial Intention for Sponsored Search Advertising",
    author = "Liu, Tongtong  and
      Wang, Zhaohui  and
      Qin, Meiyue  and
      Lu, Zenghui  and
      Chen, Xudong  and
      Yang, Yuekui  and
      Shu, Peng",
    editor = "Christodoulopoulos, Christos  and
      Chakraborty, Tanmoy  and
      Rose, Carolyn  and
      Peng, Violet",
    booktitle = "Proceedings of the 2025 Conference on Empirical Methods in Natural Language Processing",
    month = nov,
    year = "2025",
    address = "Suzhou, China",
    publisher = "Association for Computational Linguistics",
    url = "https://aclanthology.org/2025.emnlp-main.1473/",
    doi = "10.18653/v1/2025.emnlp-main.1473",
    pages = "28936--28948",
    ISBN = "979-8-89176-332-6"
}

@inproceedings{qin2023queryrewriting,
  title        = {Query Rewriting for Retrieval‑Augmented Large Language Models},
  author       = {Qin, Le and others},
  booktitle    = {Proceedings of the 2023 Conference on Empirical Methods in Natural Language Processing (EMNLP)},
  year         = {2023},
  pages        = {5313–5326},
  url          = {https://aclanthology.org/2023.emnlp-main.322},
  note         = {This work integrates web‑search engines as the retriever in a RAG pipeline.}
}

@article{xie2024_weknowrag,
  title        = {WeKnow‑RAG: An Adaptive Approach for Retrieval‑Augmented Generation Integrating Web Search and Knowledge Graphs},
  author       = {Xie, Weijian and Liang, Xuefeng and Liu, Yuhui and Ni, Kaihua and Cheng, Hong and Hu, Zetian},
  journal      = {arXiv preprint arXiv:2408.07611},
  year         = {2024},
  note         = {Combines web search with knowledge graphs in a RAG system to improve factuality.}
}

@misc{openai_web_search_tool,
  title        = {Web Search — OpenAI API Documentation},
  author       = {OpenAI},
  year         = {2025},
  howpublished = {\url{https://platform.openai.com/docs/guides/tools/web-search?api-mode=chat}},
  note         = {Accessed: 2025-11-22}
}

@misc{anthropic_claude_web_search,
  title        = {Web Search Tool — Claude Docs},
  author       = {Anthropic},
  year         = {2025},
  howpublished = {\url{https://docs.claude.com/en/docs/agents-and-tools/tool-use/web-search-tool}},
  note         = {Accessed: 2025-11-22}
}

@misc{cheng2024dateddatatracingknowledge,
      title={Dated Data: Tracing Knowledge Cutoffs in Large Language Models}, 
      author={Jeffrey Cheng and Marc Marone and Orion Weller and Dawn Lawrie and Daniel Khashabi and Benjamin Van Durme},
      year={2024},
      eprint={2403.12958},
      archivePrefix={arXiv},
      primaryClass={cs.CL},
      url={https://arxiv.org/abs/2403.12958}, 
}

@misc{kale2025knowrlteachinglanguagemodels,
      title={KnowRL: Teaching Language Models to Know What They Know}, 
      author={Sahil Kale and Devendra Singh Dhami},
      year={2025},
      eprint={2510.11407},
      archivePrefix={arXiv},
      primaryClass={cs.CL},
      url={https://arxiv.org/abs/2510.11407}, 
}

@inproceedings{qiu-etal-2024-large,
    title = "Are Large Language Model Temporally Grounded?",
    author = "Qiu, Yifu  and
      Zhao, Zheng  and
      Ziser, Yftah  and
      Korhonen, Anna  and
      Ponti, Edoardo  and
      Cohen, Shay",
    editor = "Duh, Kevin  and
      Gomez, Helena  and
      Bethard, Steven",
    booktitle = "Proceedings of the 2024 Conference of the North American Chapter of the Association for Computational Linguistics: Human Language Technologies (Volume 1: Long Papers)",
    month = jun,
    year = "2024",
    address = "Mexico City, Mexico",
    publisher = "Association for Computational Linguistics",
    url = "https://aclanthology.org/2024.naacl-long.391/",
    doi = "10.18653/v1/2024.naacl-long.391",
    pages = "7064--7083"
}

@inproceedings{zhang-etal-2025-mrag,
    title = "{MRAG}: A Modular Retrieval Framework for Time-Sensitive Question Answering",
    author = "Zhang, Siyue  and
      Xue, Yuxiang  and
      Zhang, Yiming  and
      Wu, Xiaobao  and
      Luu, Anh Tuan  and
      Zhao, Chen",
    editor = "Christodoulopoulos, Christos  and
      Chakraborty, Tanmoy  and
      Rose, Carolyn  and
      Peng, Violet",
    booktitle = "Findings of the Association for Computational Linguistics: EMNLP 2025",
    month = nov,
    year = "2025",
    address = "Suzhou, China",
    publisher = "Association for Computational Linguistics",
    url = "https://aclanthology.org/2025.findings-emnlp.167/",
    doi = "10.18653/v1/2025.findings-emnlp.167",
    pages = "3080--3118",
    ISBN = "979-8-89176-335-7"
}

@article{chen2021timeqa,
  title        = {A Dataset for Answering Time-Sensitive Questions},
  author       = {Chen, Wenhu and Wang, Xinyi and Wang, William Yang},
  journal      = {arXiv preprint arXiv:2108.06314},
  year         = {2021},
  url          = {https://arxiv.org/abs/2108.06314}
}

@inproceedings{zhang2021situatedqa,
  title        = {SituatedQA: Incorporating Extra-Linguistic Contexts into QA},
  author       = {Zhang, Michael J.Q. and Choi, Eunsol},
  booktitle    = {Proceedings of the 2021 Conference on Empirical Methods in Natural Language Processing (EMNLP)},
  pages        = {7371--7387},
  year         = {2021},
  publisher    = {Association for Computational Linguistics},
  doi          = {10.18653/v1/2021.emnlp-main.586},
  url          = {https://aclanthology.org/2021.emnlp-main.586.pdf}
}

@ARTICLE{10.3389/frai.2024.1341697,
  
AUTHOR={Quelle, Dorian  and Bovet, Alexandre },
         
TITLE={The perils and promises of fact-checking with large language models},
        
JOURNAL={Frontiers in Artificial Intelligence},
        
VOLUME={Volume 7 - 2024},

YEAR={2024},

URL={https://www.frontiersin.org/journals/artificial-intelligence/articles/10.3389/frai.2024.1341697},

DOI={10.3389/frai.2024.1341697},

ISSN={2624-8212}}

@inproceedings{Dai2024CocktailAC,
    title = "Cocktail: A Comprehensive Information Retrieval Benchmark with {LLM}-Generated Documents Integration",
    author = "Dai, Sunhao  and
      Liu, Weihao  and
      Zhou, Yuqi  and
      Pang, Liang  and
      Ruan, Rongju  and
      Wang, Gang  and
      Dong, Zhenhua  and
      Xu, Jun  and
      Wen, Ji-Rong",
    editor = "Ku, Lun-Wei  and
      Martins, Andre  and
      Srikumar, Vivek",
    booktitle = "Findings of the Association for Computational Linguistics: ACL 2024",
    month = aug,
    year = "2024",
    address = "Bangkok, Thailand",
    publisher = "Association for Computational Linguistics",
    url = "https://aclanthology.org/2024.findings-acl.421/",
    doi = "10.18653/v1/2024.findings-acl.421",
    pages = "7052--7074"
}

@inproceedings{Li2023LLatrievalLL,
    title = "{LL}atrieval: {LLM}-Verified Retrieval for Verifiable Generation",
    author = "Li, Xiaonan  and
      Zhu, Changtai  and
      Li, Linyang  and
      Yin, Zhangyue  and
      Sun, Tianxiang  and
      Qiu, Xipeng",
    editor = "Duh, Kevin  and
      Gomez, Helena  and
      Bethard, Steven",
    booktitle = "Proceedings of the 2024 Conference of the North American Chapter of the Association for Computational Linguistics: Human Language Technologies (Volume 1: Long Papers)",
    month = jun,
    year = "2024",
    address = "Mexico City, Mexico",
    publisher = "Association for Computational Linguistics",
    url = "https://aclanthology.org/2024.naacl-long.305/",
    doi = "10.18653/v1/2024.naacl-long.305",
    pages = "5453--5471"
}

@inproceedings{Shen2023LLMsAS,
    title = "Retrieval-Augmented Retrieval: Large Language Models are Strong Zero-Shot Retriever",
    author = "Shen, Tao  and
      Long, Guodong  and
      Geng, Xiubo  and
      Tao, Chongyang  and
      Lei, Yibin  and
      Zhou, Tianyi  and
      Blumenstein, Michael  and
      Jiang, Daxin",
    editor = "Ku, Lun-Wei  and
      Martins, Andre  and
      Srikumar, Vivek",
    booktitle = "Findings of the Association for Computational Linguistics: ACL 2024",
    month = aug,
    year = "2024",
    address = "Bangkok, Thailand",
    publisher = "Association for Computational Linguistics",
    url = "https://aclanthology.org/2024.findings-acl.943/",
    doi = "10.18653/v1/2024.findings-acl.943",
    pages = "15933--15946"
}

@misc{Sun2024LMORTLL,
      title={LLM-Oriented Retrieval Tuner}, 
      author={Si Sun and Hanqing Zhang and Zhiyuan Liu and Jie Bao and Dawei Song},
      year={2024},
      eprint={2403.01999},
      archivePrefix={arXiv},
      primaryClass={cs.CL},
      url={https://arxiv.org/abs/2403.01999}, 
}

@misc{Thakur2021BEIRAH,
      title={BEIR: A Heterogenous Benchmark for Zero-shot Evaluation of Information Retrieval Models}, 
      author={Nandan Thakur and Nils Reimers and Andreas Rücklé and Abhishek Srivastava and Iryna Gurevych},
      year={2021},
      eprint={2104.08663},
      archivePrefix={arXiv},
      primaryClass={cs.IR},
      url={https://arxiv.org/abs/2104.08663}, 
}

@inproceedings{Wang2024BenchmarkingLL,
author = {Chen, Jiawei and Lin, Hongyu and Han, Xianpei and Sun, Le},
title = {Benchmarking large language models in retrieval-augmented generation},
year = {2024},
isbn = {978-1-57735-887-9},
publisher = {AAAI Press},
url = {https://doi.org/10.1609/aaai.v38i16.29728},
doi = {10.1609/aaai.v38i16.29728},
booktitle = {Proceedings of the Thirty-Eighth AAAI Conference on Artificial Intelligence and Thirty-Sixth Conference on Innovative Applications of Artificial Intelligence and Fourteenth Symposium on Educational Advances in Artificial Intelligence},
articleno = {1980},
numpages = {9},
series = {AAAI'24/IAAI'24/EAAI'24}
}

@misc{kale2025liemeknowledgegraphs,
      title={Lie to Me: Knowledge Graphs for Robust Hallucination Self-Detection in LLMs}, 
      author={Sahil Kale and Antonio Luca Alfeo},
      year={2025},
      eprint={2512.23547},
      archivePrefix={arXiv},
      primaryClass={cs.CL},
      url={https://arxiv.org/abs/2512.23547}, 
}

\appendix

\section{Dataset Curation Details}
\label{app:dataset-curation}

\begin{table*}[t]
\centering
\caption{Examples of retrieval failure types}
\label{tab:error_examples}
\resizebox{\textwidth}{!}{
\begin{tblr}{
  cells = {t},
  colspec = {
    p{0.08\textwidth}
    p{0.26\textwidth}
    p{0.14\textwidth}
    p{0.26\textwidth}
    p{0.26\textwidth}
  },
  hline{1,2,5} = {-}{},
}
Error Type &
Question &
LLM Search Query &
LLM Answer &
Gold Answer \& Reference \\

QF &
Who was the chief executive officer of Open Knowledge Foundation as of 2016? &
OKF CEO &
Renata Avila \newline
[\url{en.wikipedia.org/wiki/Open_Knowledge_Foundation}] &
Pavel Richter \newline
[\url{blog.okfn.org/2015/04/29/open-knowledge-appoints-pavel-richter-as-new-ceo/}] \\

SF &
Who was the president of Mozambique in 1996? &
Mozambique president in 1996 &
Samora Moises Machel \newline
[\url{nytimes.com/1975/06/25/archives/mozambiques-first-president-samora-moises-machel.html}] &
Joaquim Chissano \newline
[\url{en.wikipedia.org/wiki/List_of_presidents_of_Mozambique}] \\

EF &
Where was the latest Summer Olympic Games held? &
Summer Olympics host cities &
Tokyo \newline
[\url{en.wikipedia.org/wiki/List\_of\_Olympic\_Games\_host\_cities}] &
Paris (2024) \newline
[\url{en.wikipedia.org/wiki/List\_of\_Olympic\_Games\_host\_cities}] \\

\end{tblr}
}
\end{table*}

\subsection{Static Split}
\label{app:static-curation}
The static split was constructed through a combination of deterministic temporal filtering, manual query validation and de-duplication, and an independent annotation-based quality check. The objective of the curation process was to obtain factual queries that are temporally anchored, fully
specified, unambiguous, and answerable through a single factual lookup, while avoiding examples requiring multi-step reasoning or extensive information synthesis.

\subsubsection{Temporal and Factual Filtering}
We begin with the 1.24k examples in the TempRAGEval test set. The original dataset provides two fields used for temporal filtering: \texttt{temporal relation} and \texttt{key date}. We retain examples whose \texttt{temporal relation} belongs to set of: \texttt{as of}, \texttt{before}, \texttt{between}, \texttt{by}, \texttt{in}, \texttt{until}, \texttt{on}.

These relations identify examples containing an explicit temporal grounding of the fact being queried. We then require the corresponding \texttt{key date} to precede 1 February 2025, which is the earliest relevant knowledge
cutoff among the evaluated model configurations. This filtering step is deterministic and does not involve model judgments. Its purpose is to ensure that the retained examples have an explicit temporal grounding and that the corresponding fact predates the knowledge
cutoff used in the experiments. Following this process, the initial 1,248 samples were reduced to 952.

\subsubsection{Manual Validation and De-duplication}
The temporally filtered candidate pool was manually reviewed using predefined inclusion and exclusion criteria. The review focused on four properties: uniqueness, completeness, factual specificity, and single-query
answerability. The first validation pass was performed by the authors.

\begin{itemize}
    \item \textit{Duplicate removal:} Duplicate and near-duplicate queries were removed by examining repeated question formulations and repeated underlying factual targets. Examples asking for the same fact with only superficial wording differences were treated as duplicates when retaining both would not provide a meaningfully different evaluation case.
    \item \textit{Completeness and specificity:} A query was retained only if all information required to identify the intended fact and construct a complete web-search query was present in the question. Queries were excluded when they contained unresolved references, omitted essential entities, lacked a required temporal or contextual qualifier, or otherwise required the annotator to infer information that was necessary to identify the intended answer.
    \item \textit{Single-query answerability:} We further excluded examples requiring multi-step reasoning, compositional search, or synthesis across multiple independent facts. An example was considered suitable when a "single complete and effective web-search query can be constructed directly from the question and the resulting search can lead to the requested factual answer without requiring a sequence of dependent searches or substantial reasoning over multiple sources."
    \item \textit{Ambiguity and answer uniqueness:} Examples were removed when multiple plausible answers remained after considering the temporal context and wording of the question, or when the intended answer could not be determined unambiguously from the available information.
    \item \textit{Semantic answer variants:} We retain the semantic answer variants provided in the original TempRAGEval \texttt{answer} field.
\end{itemize}

This process resulted in 783 samples marked for final inclusion.

\subsubsection{Independent Annotation and Dataset Validation}
\label{app:static-curation-validation}
To assess the consistency of the manual curation criteria, two independent annotators, both Master's students in computer science, reviewed the same random sample of 200 candidate queries. The annotators were provided with the
following decision task:

\begin{quote}
Given the query, determine whether the example is complete and unambiguous, contains all information required to construct an effective web-search query, and has a directly retrievable factual answer using such a query. You may use Google/Bing search manually to verify these criteria. Aslo verify that the listed answer variants are valid and complete for the corresponding query. Mark an example as 1 if all criteria are satisfied and 0 otherwise, and provide a comment explaining exclusion decisions or missing answer variations. 
\end{quote}

The annotators independently assessed the examples without consulting each other. For each independently reviewed example, annotators marked the example as suitable only when all stated criteria were satisfied. Disagreements were subsequently resolved through discussion and consensus. No cases regarding missing semantic answer variants were identified during the manual validation.

Across the 200 independently reviewed examples, the inter-annotator agreement was Cohen's $\kappa=0.86$. The high agreement indicates substantial consistency in applying the predefined inclusion criteria. The 200-example review was used as a quality-control assessment of the curation protocol rather than as a replacement for the full manual curation process. Following this validation, the 783 factual queries generated as above were included in the static split.

\subsection{Dynamic Split}
\label{app:dynamic-curation}
The dynamic split was constructed from the final static split through candidate selection, temporally neutral rewriting, post-cutoff answer validation, and semantic answer verification. The objective was to obtain queries for which the information need is genuinely current, the answer changed after the knowledge cutoff, and yet the resulting question remains a direct factual lookup.

\subsubsection{Candidate Selection}
Two independent annotators (Master's students in computer science) reviewed all 783 static queries to identify examples suitable for conversion into dynamic queries. Annotators were instructed as:

\begin{quote}
Retain an example only if (i) the queried answer concerned a
role, office, leadership position, or another attribute that has a strong possibility of changing over time, and (ii) the original question can be rewritten into a present-tense question without changing its underlying meaning. Mark such questions as 1, and 0 for others with clarifying comments for each.
\end{quote}

The annotators independently assessed all candidates. Inter-annotator agreement was Cohen's $\kappa=0.84$. Disagreements were resolved through discussion and consensus, resulting in 357 candidate queries.

\subsubsection{Temporally Neutral Rewriting}
The 357 candidates were rewritten to express current information needs using GPT-5.4-mini. The rewriting prompt was as follows: 

\begin{quote}
    Remove explicit historical temporal anchors while preserving the identity of the queried entity, the information being requested, and the original semantic intent. Do not introduce new information or change the scope of the question. Where appropriate, use temporal expressions such as "currently" or "as of now" to indicate that the query asks for the present value without specifying a particular date. Provide the updated query only.
\end{quote}

The rewriting was performed only to generate candidate formulations; final dataset inclusion was determined through subsequent human validation.

\subsubsection{Post-cutoff Validation and Answer Verification}
Each rewritten query was manually checked against trusted sources, including Wikipedia revision histories and reputable news sources, to verify that the answer had changed after the knowledge cutoff used in the experiments, and this answer was recorded in the dataset. Queries whose answers had not changed, or for which the post-cutoff change could not be reliably established, were removed. This process resulted in 288 queries.

For each retained query, the current answer was recorded together with one or more supporting sources. Semantic answer variants were then generated using GPT-5.4-mini with the prompt: 

\begin{quote}
    Provide a list of semantic variations for the given terms, only in terms of writing style or formatting. Do not change any content. Include only variants that express exactly the same factual answer. Do not add plausible, inferred, or partially correct alternatives.
\end{quote}

A separate annotation pass was also conducted with two separate graduate level annotators to verify query quality and answer-set completeness using the same criteria as described in Section \ref{app:static-curation-validation} the static split. Inter-annotator agreement for this validation was Cohen's $\kappa=0.82$. Disagreements were resolved through discussion and consensus. Following this process, the final dynamic split included all 288 queries, each associated with a temporally verified current answer, supporting sources, and a curated set of acceptable semantic answer variants.

\section{Evaluation Protocol}
\label{app:eval-prompts}

All models were evaluated using the same task-level instructions for the static and dynamic splits. 

For each query, the model was instructed to provide a concise factual answer and, for the static evaluation, a self-assessed confidence score between 0 and
1. The following instruction was used:

\begin{quote}
Answer the following factual question as accurately and concisely as possible. Provide only the answer and your confidence in the answer. Confidence should be a number between 0 and 1, where 0 means you are
completely uncertain and 1 means you are completely certain.

Question: [QUERY]

Return your response in the following format:

Answer: [answer]
Confidence: [number between 0 and 1]
\end{quote}

The confidence value was parsed as a numeric value in $[0,1]$ and used only for the calibration and retrieval-selectivity analyses. No confidence value was used in determining answer correctness. Final answers were extracted from the model response and evaluated for exact match against the canonical answer or any validated semantic answer variants associated with each query. Before matching, answers were normalized. No manual correction of model outputs was performed during evaluation.

\section{Failure Taxonomy Annotation}
\label{app:failure-annotation}

We manually classify failed retrieval cases to identify the stage of the retrieval pipeline responsible for the observed error. A failure case is defined as a query for which the model invoked web search but the final answer did not match the canonical answer or any validated semantic answer variant. Each case is assigned exactly one of three mutually exclusive categories: query formulation failure (QF), source quality failure (SF), or evidence integration failure (EF).

\subsubsection{Annotation Procedure}

Two independent annotators (Master's students in computer science) reviewed the failure cases using the predefined decision protocol. For each case, annotators were provided with the original query, the model's issued search query, the retrieved results and snippets, the model's final answer, and the canonical and accepted semantic reference answers. Annotators were instructed to identify the earliest stage at which the information needed to produce the correct answer became unavailable or was incorrectly handled. The annotation instructions were as follows.

\begin{quote}
    Assign one of three provided failure categories to the provided example. Do not force fit a category if the failure is not captured by the descriptions, put in "NA" and mark your comments. When multiple issues are present, assign the category corresponding to the earliest failure that is sufficient to explain the incorrect answer. 
    \begin{enumerate}
    \item \textit{Query formulation:} Determine whether the issued search query adequately captures the information need expressed in the original question, including the required entities, temporal constraints, and other essential qualifiers. If the query is materially incomplete, irrelevant, or misleading, the case is labelled QF.
    
    \item \textit{Source quality:} If the query is adequately formulated, inspect the retrieved results to determine whether they contain sufficiently relevant and reliable evidence for the correct answer. Cases where the available evidence is outdated, irrelevant, unreliable, or insufficient are labelled SF.
    
    \item \textit{Evidence integration:} If both the query and retrieved evidence are adequate, determine whether the model correctly extracted, interpreted, and incorporated the relevant evidence into its final answer. Cases failing at this stage are labelled EF.
\end{enumerate}
\end{quote}

\subsubsection{Inter-Annotator Agreement}

The two annotators independently labelled the same set of failure cases before discussion, 150 random incorrect samples per model from the static split, and 75 each from the dynamic split. Inter-annotator agreement was measured using Cohen's $\kappa$. The resulting agreement was $\kappa = 0.79$, indicating high consistency in applying the taxonomy. All disagreements were subsequently reviewed jointly and resolved by consensus, and the consensus labels were used for the analysis reported in Section~\ref{sec:accuracy}.

\begin{table}[t]
\centering
\caption{Prompt sensitivity of verbalised confidence on 100 static-split queries. Values show Pearson correlation between confidence scores elicited using the two semantically equivalent prompt formulations.}
\label{tab:confidence-sensitivity}
\begin{tabular}{lcc}
\toprule
Condition & Claude Sonnet 4.6 & GPT-5 \\
\midrule
Without web search & 0.88 & 0.85 \\
With web search    & 0.86 & 0.84 \\
\bottomrule
\end{tabular}
\end{table}

\section{Confidence Analysis}
\label{app:conf}
\subsection{Prompt sensitivity for verbalised confidence elicitation}
To assess whether the observed calibration behaviour is sensitive to the wording used to elicit confidence, we conducted a prompt-sensitivity analysis on 100 randomly selected static-split queries using the two frontier models, GPT-5 and Claude Sonnet. We compare the original confidence instruction \textit{(rate your confidence in the answer between 0 and 1)} with a semantically equivalent formulation asking the model to \textit{accompany your answer with a confidence score between 0 and 1}. We evaluate the agreement between the confidence scores produced under the two prompts using Pearson correlation, separately for retrieval-disabled and retrieval-enabled conditions.

As shown in Table~\ref{tab:confidence-sensitivity}, the two prompt formulations produce highly correlated confidence scores across both models and conditions ($r=0.84$--$0.88$). The correlations remain similarly high when web search is enabled, suggesting that the reported confidence signal is reasonably stable to this change in elicitation wording. This check reduces concerns about the stability and reliability of the confidence signal used in our analysis.

\end{document}